\documentclass{article}

\usepackage{microtype}
\usepackage{graphicx}
\usepackage{subfigure}
\usepackage{booktabs} 

\usepackage{hyperref}

\usepackage[accepted]{icml2025}

\usepackage[T1]{fontenc}

\usepackage{amsmath}
\usepackage{amssymb}
\usepackage{mathtools}
\usepackage{amsthm}

\usepackage{hyperref}
\usepackage{url}

\usepackage{array}
\usepackage{microtype}
\usepackage{graphicx}
\usepackage{xcolor}
\usepackage{subfigure}
\usepackage{amsmath}

\usepackage{bm}
\usepackage{dsfont}
\usepackage{mathtools}
\usepackage{nccmath}
\usepackage{amssymb}
\usepackage{multirow}
\usepackage{booktabs}

\usepackage{varwidth}
\usepackage{pifont}
\usepackage{makecell}

\usepackage[font=small,labelfont=bf,tableposition=top]{caption}
\usepackage{wrapfig}
\usepackage[flushleft]{threeparttable}

\usepackage{varwidth}
\usepackage{pifont}
\usepackage{makecell}
\usepackage{enumitem}
\usepackage{soul}
\usepackage[normalem]{ulem} 
\usepackage{listings}

\setul{0.3ex}{0.6pt}

\usepackage[capitalize,noabbrev]{cleveref}

\usepackage{makecell}

\theoremstyle{plain}

\theoremstyle{definition}

\theoremstyle{remark}

\usepackage[textsize=tiny]{todonotes}

\usepackage{tikz}

\newcommand{\transparenttext}[2][blue!30]{
  \tikz[baseline=(textnode.base)]{
    \node[fill=#1, fill opacity=0.5, text opacity=1, inner sep=1pt] (textnode) {#2};
  }
}

\definecolor{sectioncolor}{HTML}{A91D3A}
\newcommand{\sectioncolor}{sectioncolor}

\icmltitlerunning{PDE-Controller: LLMs for Autoformalization and Reasoning of PDEs}

\begin{document}

\twocolumn[
\icmltitle{PDE-Controller: LLMs for Autoformalization and Reasoning of PDEs}

\icmlsetsymbol{equal}{*}

\begin{icmlauthorlist}
\icmlauthor{Mauricio Soroco}{equal,sfucs}
\icmlauthor{Jialin Song}{equal,sfucs}
\icmlauthor{Mengzhou Xia}{princeton}
\icmlauthor{Kye Emond}{sfumath,sfumphy}
\icmlauthor{Weiran Sun}{sfumath}
\icmlauthor{Wuyang Chen}{sfucs}
\end{icmlauthorlist}

\icmlaffiliation{princeton}{Department of Computer Science, Princeton University}
\icmlaffiliation{sfumath}{Department of Mathematics, Simon Fraser University}
\icmlaffiliation{sfumphy}{Department of Physics, Simon Fraser University}
\icmlaffiliation{sfucs}{School of Computing Science, Simon Fraser University}

\icmlcorrespondingauthor{Wuyang Chen}{wuyang@sfu.ca}

\icmlkeywords{Machine Learning, Large Language Model, PDE, AI-for-Math, Autoformalization, Reasoning, ICML}

\vskip 0.3in
]

\printAffiliationsAndNotice{\icmlEqualContribution} 

\begin{abstract}
While recent AI-for-math has made strides in \emph{pure} mathematics, areas of \emph{applied} mathematics, particularly PDEs, remain underexplored despite their significant real-world applications.
We present \textbf{PDE-Controller}, a framework that enables large language models (LLMs) to control systems governed by partial differential equations (PDEs).
Our approach enables LLMs to transform informal natural language instructions into formal specifications, and then execute reasoning and planning steps to improve the utility of PDE control.
We build a \textbf{holistic} solution comprising \emph{datasets (both human-written cases and 2 million synthetic samples), math-reasoning models, and novel evaluation metrics}, all of which require significant effort.
Our PDE-Controller significantly outperforms prompting the latest open-source and GPT models in reasoning, autoformalization, and program synthesis, achieving up to a 62\% improvement in utility gain for PDE control.
By bridging the gap between language generation and PDE systems, we demonstrate the potential of LLMs in addressing complex scientific and engineering challenges.
We release all data, model checkpoints, and code at~\url{https://pde-controller.github.io/}.
\end{abstract}

\section{Introduction}

Recent advancements have significantly enhanced capabilities of Large Language Models (LLMs)
~\cite{mckinzie2025mm1,huang2023visual}.
LLMs possess pretrained common knowledge and solves daily life
tasks that require \textbf{commonsense reasoning} without domain-specific expertise.
\emph{However}, this reliance on generalized knowledge exposes significant \emph{weaknesses} in \emph{complex domains}.
LLMs struggle with precise mathematical reasoning~\cite{mirzadeh2024gsm,feng2024numerical,ahn2024large}, understanding nuanced constraints~\cite{williams2024easy}, or making decisions grounded in physical-world consequences~\cite{wang2024large,jia2024decision,cheng2024llm+}.
Addressing these limitations will require enhancing LLMs
with external tools or
domain-specific reasoning.

\begin{figure}[t!] 
\vspace{-.5em}
	\centering
	\includegraphics[width=0.38\textwidth]{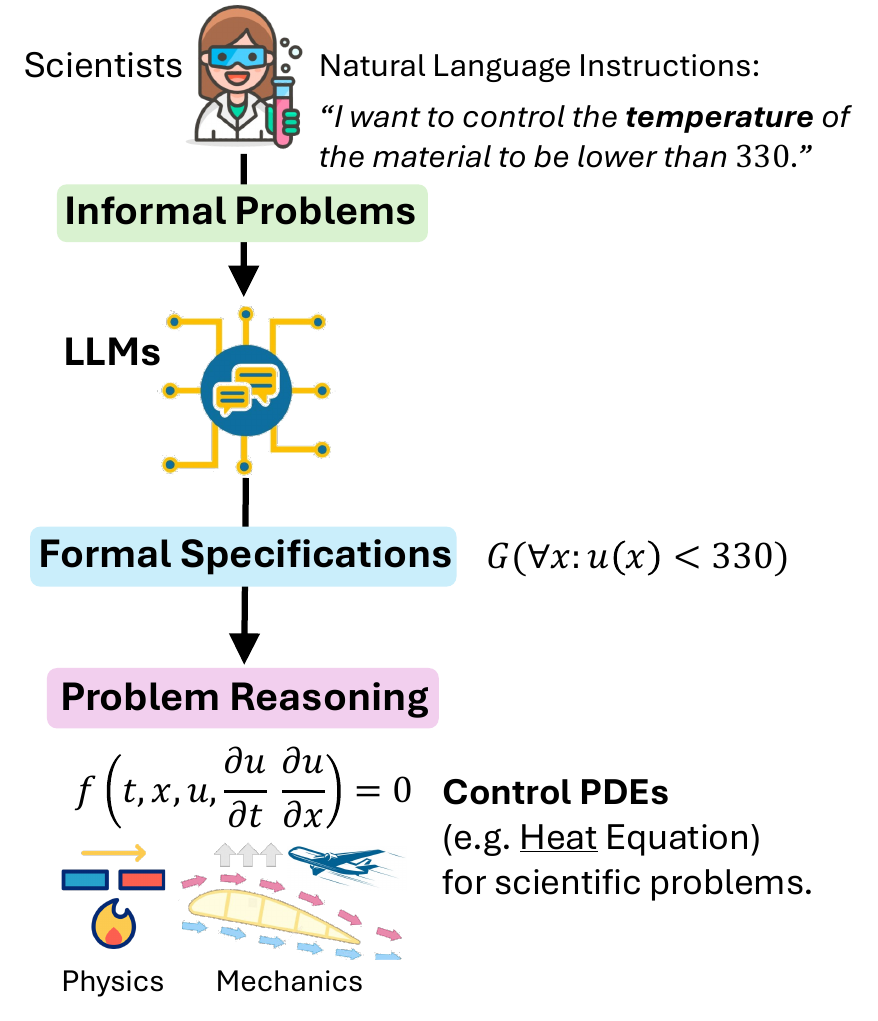}
    \vspace{-0.5em}
    \captionsetup{font=small}
    \caption{We build LLMs for automated, accelerated PDE control.}
	\label{fig:teaser}
    \vspace{-2em}
\end{figure}

Recent advancements in \textbf{AI-for-math}~\cite{lu2022survey,li2024survey} have significantly enhanced LLMs in logical, formal, and quantitative reasoning, particularly for \ul{pure mathematics} (geometry, probability, calculus, algebra, number theory, and combinatorics).
These efforts
address challenges from grade school math~\cite{cobbe2021training} to the International Mathematical Olympiad~\cite{hendrycks2021measuring,trinh2024solving}.
However, the advancement of LLMs for \textbf{applied mathematics}, such as partial differential equations (\textbf{PDEs}), remains underexplored.
Unlike pure mathematics for abstract theory,
\emph{
applied mathematics directly addresses practical challenges, bridging theory and real-world needs.
}
For example, PDEs are fundamental in modeling physical dynamics (aerospace engineering, quantum mechanics, fluid dynamics), providing a framework to understand and control systems.
Integrating LLMs into applied mathematics, particularly for \textbf{PDE control}, holds substantial potential for advancing scientific and engineering applications.

Solving PDE control problems has never been easy.
Traditional approaches
like
optimization~\cite{mcnamara2004fluid} and formal methods~\cite{alvarez2020formal}
suffer from two bottlenecks:
\emph{First}, manual PDE control
requires \ul{significant human efforts}
to understand problem descriptions and formalize into specifications 
~\cite{alvarez2020formal}.
\emph{Second}, PDE control requires \ul{highly specialized knowledge} on both coding and physics,
which are challenging even for seasoned mathematicians and engineers.
The alternative is to leverage pretrained LLMs.
However, commonsense logic captured by popular language datasets~\cite{gao2020pile,weber2024redpajama} largely deviates from scientific reasoning that requires math and physics backgrounds, leading to poor performance (Table~\ref{table:translator_coder} and~\ref{table:controller}).
Moreover, unlike conventional programming languages (Python, Java, etc.) that are rich on GitHub~\cite{chen2021evaluating}, formal languages and special libraries required by PDE control
(Fig.~\ref{fig:dataset}) are mainly used by relatively few mathematicians and engineers, resulting in limited language datasets.
Thus, our core question is:

\begin{center}
\vspace{-0.7em}
\textit{\textbf{Can LLMs control PDEs with scientific reasoning?}}
\vspace{-0.7em}
\end{center}

In this work, we aim to \textbf{advance AI-for-math for PDEs}, making open-loop PDE control automated and accessible to broad scientific practitioners with reduced efforts (Fig.~\ref{fig:teaser}).
Our framework, \textbf{PDE-Controller}, provides affirmative answers.
PDE-Controller enables autoformalization of informal PDE control problems into formal specifications and executable code, and enhances scientific reasoning by proposing novel subgoals to improve open-loop control utility.
This leads to a new methodology for integrating LLMs into scientific computing.
Our technical contributions are summarized below:
\vspace{-1em}
\begin{enumerate}[leftmargin=*]
    \setlength\itemsep{0em}
    \item \textbf{Scientific Reasoning.}
    Our PDE-Controller
    achieves up to a 62\% improvement in PDE control utility gain, compared to prompting the latest LLMs.
    This demonstrates a promising approach to future LLM-based scientific reasoning.
    Our Controller is trained via reinforcement learning from human feedback (RLHF), with rewards derived from PDE simulations labeled as win or lose.
    \item \textbf{Autoformalization and program synthesis.}
    We train LLMs via supervised fine-tuning (SFT) to automatically formalize PDE control problems, transforming informal natural language descriptions into formal specifications (over 64\% accuracy) and executable programs that integrate with external tools (over 82\% accuracy).
    \item \textbf{New Datasets.}
    We build the \emph{first} comprehensive datasets for PDE control designed for LLMs, including over \emph{2 million} samples of natural and formal language, code programs, as well as PDE control annotations.
    We also collect manually written samples by \emph{human volunteers} to evaluate LLMs in real-world scenarios.
    Our novel dataset will serve as a high-quality testbed for future research in AI for PDE reasoning.
\end{enumerate}

\section{Preliminaries}
\label{sec:preliminaries}

\subsection{Background of PDE Control}

\begin{figure}[t!] 
	\centering
	\includegraphics[width=0.48\textwidth]{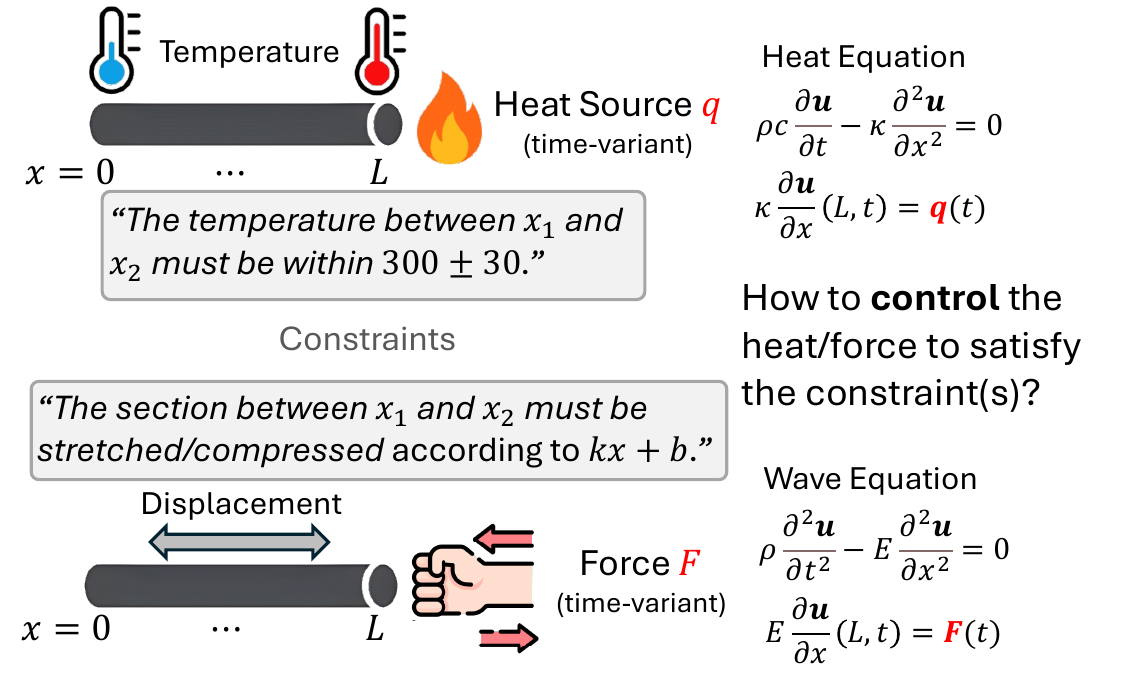}
    \vspace{-1.5em}
    \captionsetup{font=small}
    \caption{PDE control adjusts inputs (heat, force) to ensure systems (modeled by PDEs) satisfy spatiotemporal constraints.}
	\label{fig:pde_control}
    \vspace{-0.5em}
\end{figure}

Partial differential equations (PDEs) model nearly all of the physical systems and processes of interest to scientists and engineers.
PDE control involves adjusting external inputs like heat or force to guide a system governed by physical laws (PDEs) to meet specific goals or constraints.
For example, heat flow and mechanical stretching/compression in a rod are modeled and controlled by the heat and wave equations (Fig.~\ref{fig:pde_control}).
The goal is to maintain the rod's temperature or deformation within a safe range, which requires precise, time-varying control of the heat source and applied force.
This is challenging because PDEs describe complex interactions across space and time, and small changes at one point can affect the entire system.
Essentially, PDE control ensures the system behaves predictably and stays within desired limits.
We provide more details in Appendix~\ref{appendix:pdes}.

\begin{figure*}[t!]
\vspace{-0.5em}
	\centering
	\includegraphics[width=0.99\textwidth]{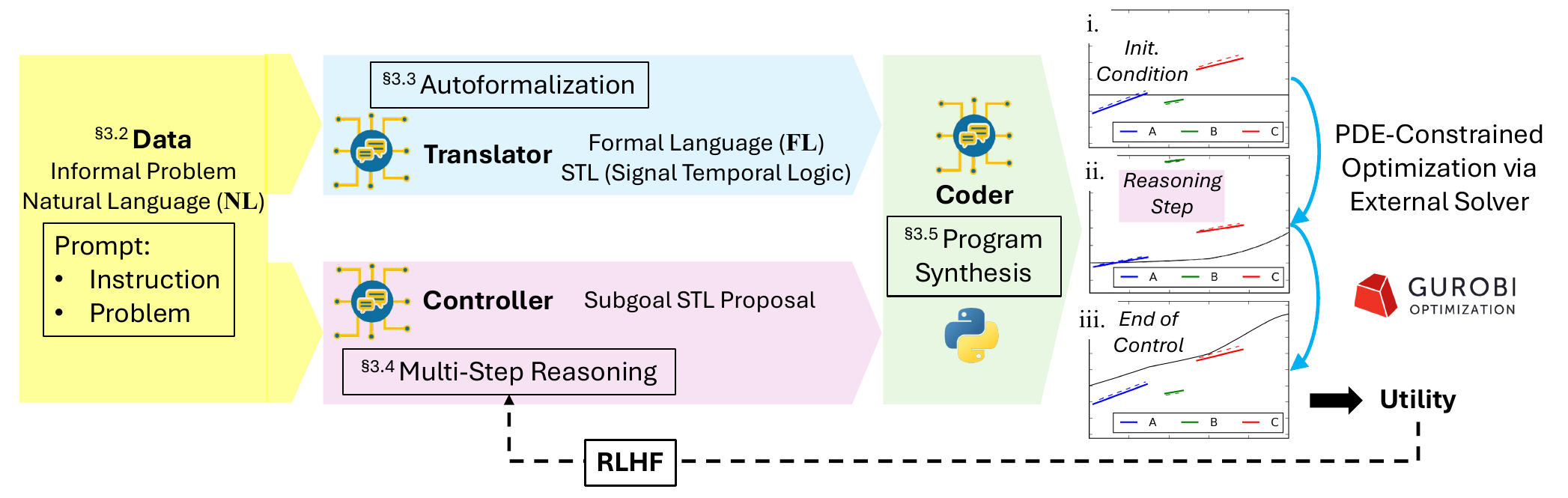}
    \vspace{-1.em}
    \captionsetup{font=small}
    \caption{Overview of our PDE-Controller framework.
     The Translator directly autoformalizes an informal PDE control problem (\transparenttext[yellow!70]{yellow}) into formal specifications with STL (\transparenttext[cyan!30]{blue}).
     The Controller proposes novel STL subgoals (\transparenttext[purple!30]{purple}). Each STL is synthesized into specialized Python programs by the Coder (\transparenttext[green!30]{green}) 
     and optimized externally (white).
     From the initial condition (i.), our PDE reasoning optimizes a subgoal (ii.) before the original problem, improving the utility at the end of control (iii.).
     We train the Controller with reinforcement learning from human feedback (RLHF).}
	\label{fig:overview}
    \vspace{-1em}
\end{figure*}

\begin{table*}[t!]
\centering
\caption{Overview of our LLM components. ``NL'': natural language. ``STL'': Signal Temporal Logic.}
\vspace{-0.5em}
\resizebox{0.95\textwidth}{!}{
\begin{tabular}{lccccc}
\toprule
\textbf{LLM} & \textbf{Purpose} & \textbf{Training Method} & \textbf{Input} & \textbf{Output} & \textbf{Evaluation Metric (s)} \\ \midrule
\makecell{Translator\\(Sec.~\ref{sec:autoformalization})} & \makecell{Autoformalize constraints from\\informal NL into formal STL} & SFT & NL & STL & \makecell{IoU: true vs.\\predicted constraints} \\ \midrule
\makecell{Controller\\(Sec.~\ref{sec:reasoning})} & \makecell{Propose subgoal STLs as intermediate\\reasoning steps to improve the final control utility} & RLHF (DPO) & NL + STL & Subgoal STL & \makecell{Success Rate (P)\\Utility Gain ($\Delta r$)} \\ \midrule
\makecell{Coder\\(Sec.~\ref{sec:program_synthesis})} & \makecell{Generate Python code for\\PDE solver (Gurobi)} & SFT & NL + STL & Python code & \makecell{Executability (\%),\\Utility RMSE (on $r(\phi))$} \\
\bottomrule
\end{tabular}
}
\label{table:dataset_formalization}
\vspace{-1em}
\end{table*}

\subsection{Formal Methods for PDE Control}
\label{sec:prelim_tradition}

\paragraph{Signal Temporal Logic.}

Following previous works, we use signal temporal logic (STL)~\cite{maler2004monitoring,alvarez2020formal} to formally represent constraints in PDE control problems:
\vspace{-0.5em}
\begin{equation}
    \phi = \mathcal{T}_{[t_{1}, t_{2}]} \left( \forall x \in [x_{1}, x_{2}], u(x) \lessgtr (ax + b) \right)
\label{eq:stl}
\end{equation}
where $\mathcal{T} \in \{\mathbf{G}, \mathbf{F}\}$ and $\lessgtr$ indicates a choice from $\{<, >, =\}$.
Specifically:
\vspace{-0.5em}
\begin{itemize}[leftmargin=*]
    \setlength\itemsep{0em}
    \item Each STL $\phi$ defines a spatiotemporal constraint on the target variable $u$ (the quantity to be controlled, like temperature, displacement).
    For simplicity, we only consider time-invariant linear constraints $ax + b$. For example: $\forall x \in [x_{1}, x_{2}], \forall t \in [t_{1}, t_{2}]$, $u(x, t) - \left( \frac{x}{2} + 300 \right) \geq -3$.
    \item $\mathbf{G}$ (``globally'') means the constraint holds \emph{during} a specified interval. $\mathbf{F}$ (``eventually'') means the constraint is satisfied \emph{at least once} during the temporal interval.
    \item Composing multiple constraints can form more complicated constraints.
\end{itemize}

Representing PDE constraints using STL can clearly and precisely express complex specifications into logical formulas and thus remove possible ambiguity in informal natural language.
In addition to binary semantics (``satisfied'' or ``unsatisfied'') defined above, STL admits continuous semantics as \textbf{utility}~\cite{kress2009temporal,donze2010robust}. 
The utility an STL $\phi$ achieves (via simulation and optimization) can be denoted as $r(\phi) \in \mathbb{R}$.
Please refer to Appendix~\ref{appendix:utility_stl} for the calculation of $r(\phi)$.

\paragraph{Problem Example: Heat Equation.}
\phantomsection\label{sec:prelim_example}

Consider a metallic rod of 100 mm. The temperature at one end of the rod is fixed at 300K, a heat source is applied to the other end. The temperature of the rod follows a heat equation. We want the temperature of the rod to be within 3K of the linear constraint $\mu(x)=\frac{x}{4}+300$ at all times between 4 and 5 seconds between 30 and 60 mm. Furthermore, the temperature should never exceed 345K at the point where the heat source is applied ($x=100$).
We can formulate this specification using the following composite STL formula:
\vspace{-1em}
\begin{gather*}
\phi=\mathbf{G}_{[4,5]}\left(\left(\forall x \in[30,60]: u(x)-\left(\frac{x}{4}+303\right)<0\right) \wedge\right. \\
\left.\quad\left(\forall x \in[30,60]: u(x)-\left(\frac{x}{4}+297\right)>0\right)\right) \wedge \\
\mathbf{G}_{[0,5]}(\forall x \in[100,100]: u(x)-345<0).
\label{eq:heat_stl}
\end{gather*}
\vspace{-2.5em}

For more examples, please read Appendix~\ref{appendix:case_study}.

\subsection{Optimization}
\label{sec:stl_optimization}

To solve the PDE control problem defined by initial conditions and STL constraints (Eqn.~\ref{eq:stl}),
the PDE is first discretized into a set of difference equations with the finite element method. Then, together with STL constraints, they are formulated into a mixed integer linear program (MILP)~\cite{sadraddini2015robust,alvarez2020formal} where the utility $r$ can be optimized via external optimizers like Gurobi~\cite{gurobi}.
After the optimization, if $r>0$, that means the system successfully satisfies $\phi$ using the control input (i.e. constraints are successfully met). This is a non-convex optimization problem.
Please see Appendix~\ref{appendix:milp_formulation} for detailed formulations.

\section{Methods}

Our core aim is to automate the formalization and reasoning of PDE control problems using large language models.

\subsection{Overview}

\paragraph{Problem Definition.}
As introduced in Sec.~\ref{sec:preliminaries},
the input of a PDE control problem is the natural language describing the initial state, system conditions, and target constraints.
The solution is a time-variant trajectory of the control input.
The final output we collect is the control utility $r(\phi) \in \mathbb{R}$.

\begin{figure*}[t!] 
	\centering
	\includegraphics[width=0.95\textwidth]{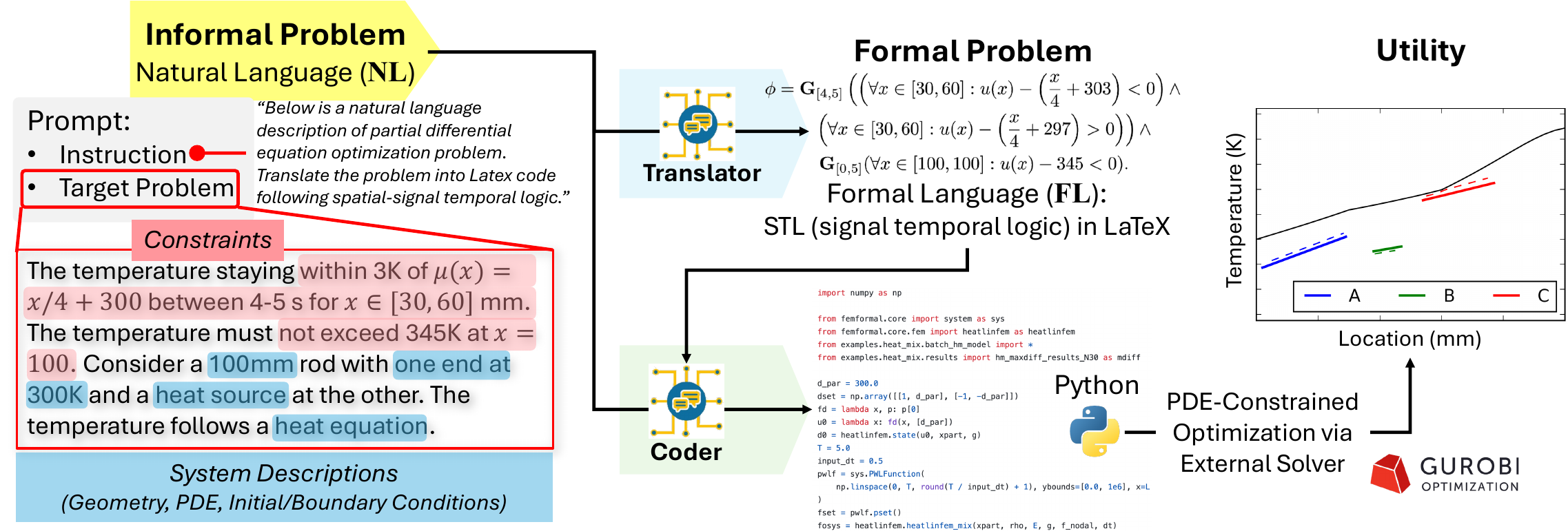}
    \vspace{-0.5em}
    \captionsetup{font=small}
    \caption{Workflow for supervised fine-tuning (SFT) of autoformalization (Translator LLM) and program synthesis (Coder LLM). Note that the utility is only used for evaluation and not used for SFT. Without reasoning, the Translator and Coder try to faithfully and directly solve the original problem.}
	\label{fig:dataset}
    \vspace{-1em}
\end{figure*}

\paragraph{Framework Overview (Fig.~\ref{fig:overview}).}
We automate the PDE control problem with the following four steps:
\vspace{-1em}
\begin{enumerate}[leftmargin=*]
    \setlength\itemsep{0em}
    \item \textbf{Input.}
    Our LLM prompt is composed of
    1) an instruction (to prompt the LLM to formalize the problem, perform reasoning, or synthesize Python programs) and
    2) the target PDE control problem.
    We build a large-scale dataset to support diverse inputs and prompts (Sec.~\ref{sec:build_data}).
    \item \textbf{Autoformalization.} Our Translator LLM will extract information about the problem constraints from the prompt, and formalize these into STLs (Sec.~\ref{sec:autoformalization}).
    \item \textbf{Reasoning.} Before directly solving the target PDE control problem, our Controller LLM will propose novel STLs as subgoals. The aim is to better solve the original PDE control problem by leveraging this subgoal as an intermediate step, i.e., we will first solve the problem defined by this subgoal which leads to a new system state, then solve the original problem (Sec.~\ref{sec:reasoning}).
    \item \textbf{Program Synthesis.} Our Coder LLM will take \textbf{both} the \ul{prompt} and the \ul{formalized STLs} as inputs, and generate Python code
    to be fed into
    the external tool (Gurobi optimizer~\cite{gurobi}) to solve the PDE control problem
    (Sec.~\ref{sec:program_synthesis}).
\end{enumerate}

\subsection{Principled Data Synthesis with Augmentations}
\label{sec:build_data}

Training LLMs for PDE problems requires diverse data, but existing math reasoning datasets~\cite{cobbe2021training,yang2024leandojo,glazer2024frontiermath} lack large-scale corpora for PDEs.
We collect key representative problems for PDE control, use them as templates, and augment them to ensure sufficient quantity and diversity.

\paragraph{Overview.}
As shown in Fig.~\ref{fig:dataset} left, each PDE control problem consists of two components: 1) constraints (\transparenttext[red!30]{red background}), 2) system descriptions (\transparenttext[cyan!30]{blue background}, including PDE, geometry, initial/boundary conditions).
We generate dataset in three steps:

\paragraph{1) Constraints: Principled Syntax Formats via STL.}
We first design and organize eligible syntax formats for generating STLs in two levels:
\vspace{-0.5em}
\begin{itemize}[leftmargin=*]
    \setlength\itemsep{0em}
    \item For each constraint, the format in Eqn.~\ref{eq:stl} leads to 6 different syntax formats ($\{\mathbf{G}, \mathbf{F}\} \times \{<, >, =\}$).
    \item We consider STLs up to 3 constraints. For 2 constraints, we connect $\phi_1$ and $\phi_2$ with logical connectives $\wedge$ or $\vee$. For 3 constraints, we consider both logical connectives and operator precedence via parentheses: $\phi_1 \vee \phi_2 \vee \phi_3, \phi_1 \wedge \phi_2 \wedge \phi_3, (\phi_1 \vee \phi_2) \wedge \phi_3, \phi_1 \vee(\phi_2 \wedge \phi_3), (\phi_1 \wedge \phi_2) \vee \phi_3, \phi_1 \wedge(\phi_2 \vee \phi_3)$. In total, STLs with 1, 2, 3 constraints will respectively result in 6, 72, 1296 unique syntax formats, i.e. 1374 in total (LaTeX in Fig.~\ref{fig:dataset} middle top).
\end{itemize}
At this moment, these STLs are still abstract, with hyperparameters like $a, b, x_1, x_2, t_1, t_2$ in Eqn.~\ref{eq:stl} not yet realized.

\paragraph{2) System Descriptions: Sampling Initial Conditions and Hyperparameters.}
To realize each STL format into a concrete problem, we need to fill numerical values for its initial conditions and hyperparameters (blue highlight in Fig.~\ref{fig:dataset} left).
We describe our sampling distributions in Appendix~\ref{appendix:parameter_ranges} for the following aspects:
\vspace{-1em}
\begin{itemize}[leftmargin=*]
    \setlength\itemsep{0em}
    \item Initial conditions: the initial temperature or displacement of the system at the beginning of the PDE simulation.
    \item Simulation domains:
     the spatial range $x_{\max}$ and temporal range $t_{\max}$.
    \item Physical properties: such as density, specific heat capacity, thermal conductivity.
    \item Coefficients: the linear parameters ($a$, $b$) and spatiotemporal ranges ($x_1$, $x_2$, $t_1$, $t_2$) for constraints.
\end{itemize}

\paragraph{3) Dataset Synthesis with Augmentations.}
We synthesize STLs into informal natural language descriptions equipped with system descriptions.
For each problem, we also synthesize ground-truth Python code
for optimization with the Gurobi solver (Fig.~\ref{fig:dataset} middle bottom).
These synthesized samples give us a one-to-one mapping from the informal problem (natural language) to its STL and Python code. Details about our dataset synthesis, are in Appendix~\ref{appendix:data_generation_rules}.
More importantly, to further promote the diversity of our informal natural language problems, we use ChatGPT (GPT4o-mini) for augmentation by rephrasing without affecting semantics.
In particular, we prompt five paraphrases from ChatGPT for every synthesized informal description.
Please see Appendix~\ref{appendix:chatgpt_augmentation} for an example of augmentation via ChatGPT.

As shown in Table~\ref{table:dataset_formalization}, in total we have $n=2.13$ million triplets of (natural language, STL, Python) samples.
We merge the training set for both heat and wave problems for the training of Translator and Coder.

\begin{table}[]
\centering
\caption{Our dataset for autoformalization and program synthesis.}
\vspace{-0.5em}
\resizebox{0.46\textwidth}{!}{
\begin{tabular}{lccc|c}
\toprule
Num. Constraints & 1 & 2 & 3 & Total Num.\\ \midrule
STLs & 6 & 72 & 1296 & 1374\\
Heat (Train) & 3840 & 45792 & 817776 & 867408 \\
Heat (Test) & 960 & 11448 & 204768 & 217176 \\
Wave (Train) & 3840 & 45504 & 795744 & 845088 \\
Wave (Test) & 960 & 11304 & 196992 & 209256\\ \bottomrule
\end{tabular}
}
\label{table:dataset_formalization}
\vspace{-1em}
\end{table}

\paragraph{Real-World Manually Written PDE Control Problems.}
\phantomsection\label{sec:manual_data}

To evaluate our LLMs on real-world problems with high variance and noise,
we collect PDE control samples designed by humans via questionnaires.
To ensure data quality, we recruit undergraduate and graduate students as participants with \ul{highly relevant backgrounds} (majoring in Math, Electic Engineering, Computer Science, Physics).
During four questionnaire sessions (one hour each), we provide comprehensive introductions to the settings of our PDE control problems, with concrete examples and interactive communication.
We collect 17 manually written heat problems and 17 wave problems.
Details of this collection and differences between our training set are shown in Appendix~\ref{appendix:manual_data}.

\subsection{Autoformalization}
\label{sec:autoformalization}

After building our dataset, we train our Translator to extract constraints from informal natural language and autoformalize into STLs.
Our LLM needs to
(1) separate constraints from system descriptions,
(2) align informal definitions and concepts to formal STL syntax,
and
(3) connect multiple constraints with correct logic operators.
This task is further complicated by context changes in different PDEs.

We leverage a pretrained MathCoder2-DeepSeekMath-7B~\cite{lu2024mathcoder2} checkpoint (MathCoder2), to fine-tune
using LoRA~\cite{hu2021lora} and supervised fine-tuning (SFT) with the cross-entropy loss.
This measures the error in predicting each token in the output formal sequence ($\widehat{\mathbf{FL}}$) given tokens in the informal input ($\mathbf{NL}$). It is defined as:
\vspace{-0.5em}
\begin{equation}
\mathcal{L}_{SFT}^{\text{translator}} = -\sum_{i=1}^n \log P\left(\widehat{\mathbf{FL}}_{i} \mid \mathbf{NL}_i, \bm{\theta}_{\text{translator}} \right)
\end{equation}

\subsection{PDE Reasoning via Controller LLM}
\label{sec:reasoning}

\subsubsection{What is Reasoning for PDE Control?}
\label{sec:pde_reasoning_definition}

Beyond autoformalization, an important question is: can LLMs show \textbf{reasoning and planning} capabilities on scientific problems like PDE control, where pretrained commonsense knowledge may not be helpful?

\paragraph{Problem Definition.}
As described in Sec.~\ref{sec:stl_optimization}, the PDE control problem is non-convex.
Directly optimizing the target anchor problem may lead to suboptimal solutions or intractability due to potentially poor initial conditions, loss landscape barriers, local minima, etc.
Inspired by recent works in robotics~\cite{lin2024clmasp,wangconformal} and AI-for-math~\cite{zawalski2022fast,zhaosubgoal,zhao2023decomposing} where subgoals are decomposed from the original problem,
we propose the following \textbf{PDE control reasoning strategy}:
\textit{The solution to a PDE control problem can be improved by decomposing it into subgoals to be optimized sequentially.}

\paragraph{PDE Control Reasoning.}
\phantomsection\label{sec:pde_reasoning_step}
We design the following reasoning steps for solving PDE control problems:
\vspace{-0.5em}
\begin{enumerate}[leftmargin=*]
    \setlength\itemsep{0em}
    \item Given a target PDE control problem $\phi$ (dubbed ``anchor''), we
    sample its STL constraints
    into a subgoal STL, $\phi^\prime$, of different spatiotemporal constraints.

    \item We directly solve the anchor problem ($\phi$) and collect its utility $r(\phi)$ and runtime cost $t$.

    \item
    We solve $\phi^\prime \rightarrow \phi$: We
    optimize $\phi^\prime$, 
    apply the system state as the new initial condition\footnote{To avoid long runtime of subgoal STLs, we set the Gurobi runtime threshold to 120 seconds for solving $\phi^\prime$.}
    solve the anchor problem $\phi$, and collect the final utility $r(\phi|\phi^\prime)$.
    \item If $r(\phi|\phi^\prime)$ $>$ $r(\phi)$ (the utility of directly solving $\phi$), we call $\phi^\prime$ a successful reasoning.
    \item We repeat the above steps multiple times to calculate the expected performance gain.
\end{enumerate}

Whether the subgoal reasoning step can be satisfied (i.e. whether or not $r(\phi^\prime) > 0$ or not) is not our concern.
All we pursue is a new initial condition of the system that can better solve the original problem. As such, the time constraints of the subgoal should apply in the period prior to the anchor constraints and within the global time frame.
Please see Fig.~\ref{fig:case_study} for concrete examples where $r(\phi|\phi^\prime)$ outperforms $r(\phi)$, and more examples in Appendix~\ref{appendix:case_study}.

\subsubsection{Building the Controller LLM.}

Having defined PDE control reasoning in Sec.~\ref{sec:pde_reasoning_definition}, our question is:
\emph{How to develop LLMs to perform PDE control reasoning and automatically decompose subgoals?}

\begin{figure}[t!] 
	\centering
	\includegraphics[width=0.475\textwidth]{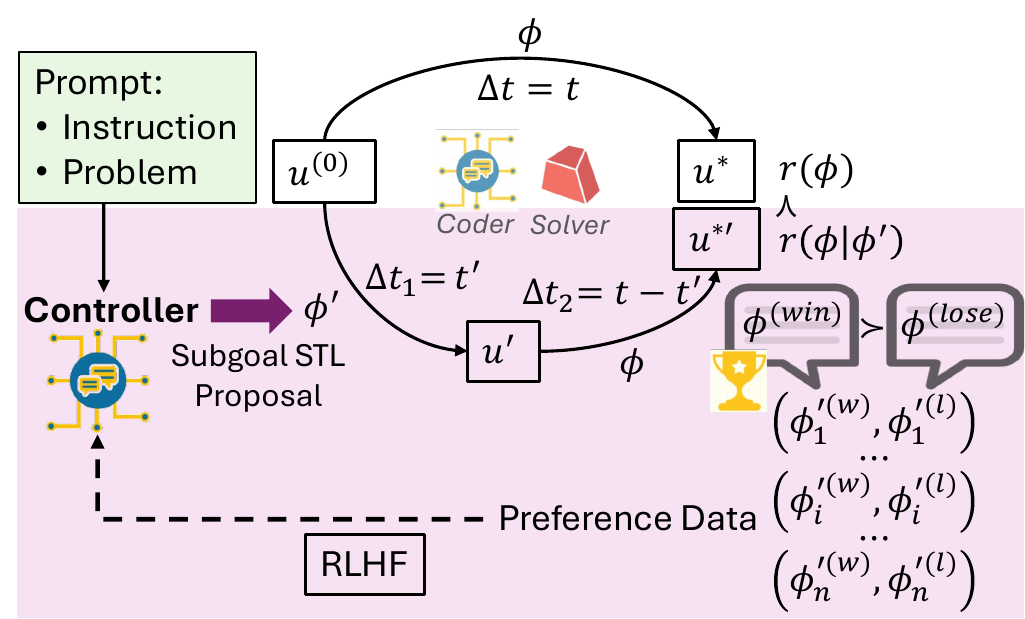}
    \vspace{-0.5em}
    \captionsetup{font=small}
    \caption{Learning PDE control reasoning via RLHF. Given the input prompt, our Controller LLM trained with preference data via reinforcement learning, will propose a subgoal STL $\phi^\prime$. From the initial condition
    $u^{(0)}$,
    the PDE system is controlled by $\phi^\prime$ to reach state
    $u^\prime$,
    and then further controlled
    by the original STL $\phi$ to reach the final state
    $u^{*\prime}$.
    We expect the utility achieved via this reasoning, $r(\phi|\phi^\prime)$, to outperform $r(\phi)$ achieved by directly solving $\phi$.
    }
	\label{fig:reasoning}
    \vspace{-1em}
\end{figure}

In this section, we explain how to train a Controller LLM via reinforcement learning with human feedback (RLHF) that can perform PDE control reasoning. We illustrate our training strategy in Fig.~\ref{fig:reasoning}.

\paragraph{1) Preparing Preference Dataset.}
We first build a dataset of paired STLs $(\phi^{\prime(w)}, \phi^{\prime(l)})$ as win-lose pairs, where $\phi^{\prime(w)}$ is preferred over $\phi^{\prime(l)}$. This dataset is used to train our Controller LLM via RLHF.
To build this dataset, given an anchor (target) STL $\phi$, we randomly sample $\phi^\prime$ based on $\phi$, solve both, and collect their utilities.

It is up to human preference to determine which STL is favored.
In this work, we
prefer
$r(\phi|\phi^{\prime(w)}) > r(\phi) > \phi^{\prime(l)}$.
In total, we collect 10772 pairs of win-lose STLs.

\paragraph{2) Fine-tuning Controller LLM with RLHF.} 

During training, our Controller loads the Translator's pretrained checkpoint (inheriting the ability to faithfully formalize natural language into STLs).
We train Controller with Direct Preference Optimization (DPO)~\cite{rafailov2024direct}.
Unlike the Translator's direct formalization into true STLs, we prompt our Controller to make modifications in proposing new STL variants.
We leverage win-lose pairs in our preference dataset as feedback to fine-tune the generative distribution of the Controller to shift to the preferred STL:
\vspace{-1em}

{\small
\begin{equation}
\begin{aligned}
\mathcal{L}_{DPO}^{\text{controller}}
& =
-\mathbb{E}_{\left(\mathbf{NL}, \phi^{\prime(w)}, \phi^{\prime(l)}\right)}
\Bigg[\log \sigma\Bigg(
\beta \log \frac{P\left(\phi^{\prime(w)} \mid \mathbf{NL}\right)}
{P_{\mathrm{ref}}\left(\phi^{\prime(w)} \mid \mathbf{NL}\right)} \\
&- \beta \log \frac{P\left(\phi^{\prime(l)} \mid \mathbf{NL}\right)}
{P_{\mathrm{ref}}\left(\phi^{\prime(l)} \mid \mathbf{NL}\right)}
\Bigg) \Bigg] 
+ \lambda \log P\left(\widehat{\mathbf{FL}} | \mathbf{NL} \right).
\label{eq:dpo_stl}
\end{aligned}
\end{equation}
}
$\sigma$ is the sigmoid function.
We load the pretrained checkpoint of our Translator LLM as the frozen reference model (``ref''), which serves as the KL divergence target in DPO training.
$P$ and $P_{\mathrm{ref}}$ indicate the generative probability of Controller and Translator respectively ($\bm{\theta}$ omitted for simplicity).
$\beta$ controls the deviation of the Controller from the reference Translator.
To further avoid the degradation of Controller's generation, we regularize it with the SFT loss, with a weight of $\lambda$~\cite{pang2024iterativereasoningpreferenceoptimization},
which we also found prevented overfitting compared to the DPO loss alone.

\subsection{Program Synthesis}
\label{sec:program_synthesis}

Finally, taking both the prompt (instruction plus the target PDE control problem) and the formalized STL as inputs, we train a Coder LLM with supervised fine-tuning (SFT) to synthesize Python programs provided to the PDE simulator and Gurobi optimizer to solve the PDE control problem.
\begin{equation}
\mathcal{L}_{SFT}^{\text{coder}}=-\sum_{i=1}^n \log P\left(\widehat{\mathbf{Code}}_{i} \mid \widehat{\mathbf{FL}}_{i}, \mathbf{NL}_i, \bm{\theta}_{\text{coder}} \right)
\end{equation}
Similar to Translator, we merge the training set for both heat and wave problems for the Coder's fine-tuning.

\section{Experiments}

We study 1D heat and wave problems as pioneering showcases. All our models are fine-tuned from MathCoder2, and we compare against few-shot evaluations of MathCoder2, GPT 4o, and GPT o1-mini~\cite{achiam2023gpt}.
Please read Appendix~\ref{appendix:training_details} for model and training details.

\vspace{-1em}
\subsection{Accurate Autoformalization and Program Synthesis}

We first evaluate the performance of our Translator for \mbox{autoformalization} and Coder for program synthesis.

\vspace{-1em}
\paragraph{Evaluation Metrics.}

To progressively evaluate the performance of our Translator and Coder LLMs in a \emph{decoupled and fine-grained} manner, we propose to leverage multiple metrics for different purposes, as summarized in Table~\ref{table:metrics}.
Among all metrics, the ``Utility RMSE'' (lower the better) is the most important final performance. However, it is important to emphasize that we can only calculate RMSE for executable Python programs. Consequently, if we observe a coder achieving a low RMSE but also low executability, it still implies poor quality.

\begin{table}[h!]
\centering
\caption{Metrics for the evaluation of autoformalization (Translator) and program synthesis (Coder).}
\vspace{-1em}
\resizebox{0.48\textwidth}{!}{
\begin{tabular}{p{0.52\linewidth} | p{0.48\linewidth}}
\toprule
Purpose & Metric \\ \midrule
LLM-generated STL ($\hat{\phi}$) could be valid, but there might be semantic mistakes\footnotemark. & \textbf{IoU}: Intersection over union (satisfying areas) between the true STL, $\phi$, and the LLM's generation, $\hat{\phi}$.\\ \midrule
Python programs generated by Coder may not be runnable due to bugs. & \textbf{Executability}: The ratio of executable programs to the total. \emph{This doesn't ensure the executed result (utility $r$) is correct.}\\ \midrule
Compare the final PDE control utility $r(\hat{\phi})$
to true utlity $r(\phi)$.
& \textbf{Utility RMSE}: Relative mean square error on utility. \\
\bottomrule
\end{tabular}
}
\label{table:metrics}
\vspace{-1em}
\end{table}
\footnotetext[2]{For example, switching from $\geq$ to $\leq$ will lead to a completely different constraint, but the STL itself is still a valid logic.}

\paragraph{Results\footnotemark.}

\footnotetext[3]{
Remarks for Table~\ref{table:translator_coder} and~\ref{table:translator_coder_human}:
1) Since IoU $\in [0, 1]$, we set the IoU of any invalid STL as 0.
2) To isolate the evaluation of program synthesis without being distracted by possibly generated bad STLs,
when calculating the executability and utility RMSE, LLMs are provided with true STLs in their prompts, instead of LLM-generated STLs.
Table~\ref{table:translator_coder_its_own_sstl_synthetic} and~\ref{table:translator_coder_its_own_sstl_human} show end-to-end results but cannot decouple the quality of program synthesis.
4) The utility RMSE is only calculated for executable Python programs.
}

The autoformalization can be evaluated using the intersection over union (IoU) between the predicted and target STLs (constraints).
The code generation should aim for
high executability and low utility RMSE \ul{simultaneously\footnote{Otherwise, the code generation might trivially generate for example a runnable ``\texttt{return 0}'', which is obviously incorrect; or it might get lucky and generate the correct code for only a very limited set of problems.}}.
As
in Table~\ref{table:translator_coder},
our Translator and Coder achieve the best across all
metrics with low
deviations, indicating strong and reliable autoformalization and program synthesis.

We further evaluate 
manually written problems.
As shown in Table~\ref{table:translator_coder_human},
the autoformalization may produce STLs with worse quality (lower IoU), suffering from noisy and unstructured texts written by humans.
Our Translator and Coder generally outperform other baselines, with the accuracy of autoformalization over 64\% (STL's IoU) and program synthesis over 82\% (code executability).
Importantly, despite the low utility RMSE of GPT o1-mini, since it suffers from poor executability (only 39.22\%), its RMSE cannot faithfully characterize its stability in the real world.

\begin{table}[h!]
\vspace{-0.5em}
\centering
\caption{Autoformalization and program synthesis.
Deviations over 3 seeds are in parentheses.
\textbf{Bold} indicates the best, \ul{underline} denotes the runner-up.
\vspace{-0.5em}
}
\resizebox{0.48\textwidth}{!}{
\addtolength{\tabcolsep}{-0.3em}
\begin{tabular}{ccccc}
\toprule
PDE & Model & \begin{tabular}{@{}c@{}}IoU ($\uparrow$)\\(Translator)\end{tabular} & \begin{tabular}{@{}c@{}}Executability ($\uparrow$)\\(Coder)\end{tabular} & \begin{tabular}{@{}c@{}}\textbf{Utility}\\\textbf{RMSE} ($\downarrow$)\end{tabular} \\ \midrule
\multirow{4}{*}{Heat} & Ours & \textbf{0.992} (0.07) & \textbf{0.9978} (0.0015) & \textbf{0.0173} (0.0065) \\
                     & MathCoder2 & \ul{0.772} (0.35) & 0.9592 (0.0166) & 0.2058 (0.0672) \\
                     & GPT (4o) & - & 0.5807 (0.3244) & 0.0445 (0.0437) \\
                     & GPT (o1-mini) & - & \ul{0.3561} (0.2857) & \ul{0.0898} (0.0165) \\
\midrule
\multirow{4}{*}{Wave} & Ours & \textbf{0.992} (0.07) & \textbf{0.9620} (0.0098) & \textbf{0.0076} (0.0011) \\
                     & MathCoder2 &  \ul{0.772} (0.35) & \ul{0.9340} (0.0242) & 0.1089 (0.0485) \\
                     & GPT (4o) & - & 0.6799 (0.2523) & 0.0868 (0.0500) \\
                     & GPT (o1-mini) & - & 0.4041 (0.2771) & \ul{0.0757} (0.0149) \\
\bottomrule
\end{tabular}
}
\label{table:translator_coder}
\vspace{-0.5em}
\end{table}

\begin{table}[h!]
\centering
\caption{Autoformalization and program synthesis on \ul{manually written data} (Sec.~\ref{sec:manual_data}).
Deviations over 3 seeds are in parentheses.
\textbf{Bold} indicates the best, \ul{underline} denotes the runner-up.
}
\vspace{-0.5em}
\resizebox{0.48\textwidth}{!}{
\addtolength{\tabcolsep}{-0.3em}
\begin{tabular}{ccccc}
\toprule
PDE & Model & \begin{tabular}{@{}c@{}}IoU ($\uparrow$)\\(Translator)\end{tabular} & \begin{tabular}{@{}c@{}}Executability ($\uparrow$)\\(Coder)\end{tabular} & \begin{tabular}{@{}c@{}}\textbf{Utility}\\\textbf{RMSE} ($\downarrow$)\end{tabular} \\ \midrule
\multirow{4}{*}{Heat} & Ours & \textbf{0.7108} (0.0043) & \ul{0.8235} (0.0) & 2.4687 (0.0)\\
 & MathCoder2 & \ul{0.3383} (0.068) &  \textbf{0.9804} (0.0277) & \textbf{0.0004} (0.0005) \\
 & GPT (4o) & - &  0.4314 (0.3328) & 1.8555 (0.0136) \\
 & GPT (o1-mini) & - &  0.3530 (0.3050) & \ul{0.1738} (0.0207) \\
\midrule
\multirow{4}{*}{Wave} & Ours & \textbf{0.6493} (0.0) &  \textbf{1.0} (0.0) & 0.0119 (0.0) \\
 & MathCoder2 & \ul{0.1953} (0.045) &  \textbf{1.0} (0.0) & 0.0129 (0.0012) \\
 & GPT (4o) & - &  \ul{0.5882} (0.4437) & \ul{0.0105} (0.0) \\
 & GPT (o1-mini) & - &  0.3922 (0.4160) & \textbf{0.0098} (0.0) \\
\bottomrule
\end{tabular}
}
\label{table:translator_coder_human}
\end{table}

\vspace{-0.5em}
\paragraph{Generalization to Unseen STL Formats.}

Although we only collect data with no more than three STLs,
it is still possible to scale up our method to more constraints.
To evaluate our generalization to unseen STL formats, we extend our method to unseen 4-constraint STLs.
We evaluate our Translator (IoU) and Coder (executability \& utility) over 5 problems each for heat and wave problems in Table \ref{table:4-stls}, and compare against MathCoder2.
We can see that our method generalizes much better than MathCoder2.
Neither model has seen problems with 4 STL constraints.
MathCoder2 is evaluated with 2 in-context examples of 4 constraints.

\begin{table}[h!]
\centering
\caption{Test different models on \ul{unseen 4-constraint} STL formats, based on IoU, Executability, and Utility RMSE. Bold indicates the best. Deviations over 5 seeds are in parentheses.}
\resizebox{0.48\textwidth}{!}{
\addtolength{\tabcolsep}{-0.3em}
\begin{tabular}{c c c c c}
\toprule
PDE & Model & \begin{tabular}{@{}c@{}}IoU ($\uparrow$)\\(Translator)\end{tabular} & \begin{tabular}{@{}c@{}}Executability ($\uparrow$)\\(Coder)\end{tabular} & \begin{tabular}{@{}c@{}}\textbf{Utility}\\\textbf{RMSE} ($\downarrow$)\end{tabular} \\ \midrule
\multirow{2}{*}{Heat} & Ours         &\textbf{ 0.934} (0.0)      & \textbf{0.8} (0.0)     & \textbf{0.0} (0.0)   \\ 
                      & MathCoder2   & 0.8154 (0.0)     & 0.6 (0.0)     & 0.2600 (0.0) \\ \hline
\multirow{2}{*}{Wave} & Ours         & \textbf{1.0} (0.0)        & \textbf{0.8} (0.0)     & \textbf{0.1515} (0.0) \\ 
                      & MathCoder2   & 0.9690 (0.0)     & \textbf{0.8} (0.0)     & 0.2393 (0.2268) \\ \hline
\end{tabular}
}
\label{table:4-stls}
\end{table}

\paragraph{Ablation Comparisons.}

We further demonstrate the necessity of our Translator and Coder in 
Table \ref{table:performance_comparison}.
We observe the following by making comparisons:
\begin{itemize}[leftmargin=*]
    \item Our Translator has better autoformalization abilities, improving +28.5\% IoU over MathCoder2.
    \item Our Coder is robust to noisy autoformalization:
    \begin{itemize}[leftmargin=*]
        \item Given ground truth STLs, our Coder has +4\% better executability rate in generated Python than MathCoder2, and utility RMSE is 91.6\% lower (better) than MathCoder2.
        \item When switching from ground truth STL inputs to noisy Translator predictions, our Coder’s utility RMSE is only 0.57\% worse, indicating that our Coder is robust under noisy predicted STL inputs.
    \end{itemize}
     
\end{itemize}

\begin{table}[h!]
\centering
\caption{Ablation comparisons of translators and coders. Bold indicates the performance change relative to the row above.}
\resizebox{0.48\textwidth}{!}{
\addtolength{\tabcolsep}{-0.3em}
\begin{tabular}{lcc}
\toprule
Method & Metric & Performance \\ 
\midrule
Mathcoder’s Translator (Table \ref{table:translator_coder}) & \multirow{2}{*}{IoU ($\uparrow$)} & 0.772 \\ 
Our Translator (Table \ref{table:translator_coder}) &  & 0.992 \textbf{(+28.5\%)} \\ 
\midrule
Mathcoder’s Coder (Table \ref{table:translator_coder}) & \multirow{2}{*}{Executability ($\uparrow$)} & 0.9592 \\ 
Our Coder (Table \ref{table:translator_coder}) &  & 0.9978 \textbf{(+4.02\%)} \\ 
\midrule
Mathcoder’s Coder (Table \ref{table:translator_coder}) & \multirow{2}{*}{Utility RMSE ($\downarrow$)} & 0.2058 \\ 
Our Coder (Table \ref{table:translator_coder}) &  & 0.0173 \textbf{(-91.6\%)} \\ 
\midrule 
Translator STL $\to$ Coder (Table \ref{table:translator_coder_its_own_sstl_synthetic}) &  \multirow{2}{*}{Utility RMSE ($\downarrow$)} & 0.0174 \\ 
Ground Truth STL $\to$ Coder(Table \ref{table:translator_coder}) &  & 0.0173 \textbf{(-0.57\%)} \\ 
\bottomrule
\end{tabular}
}
\label{table:performance_comparison}
\end{table}

\begin{figure*}[t!]
    \centering
    \includegraphics[width=0.99\linewidth]{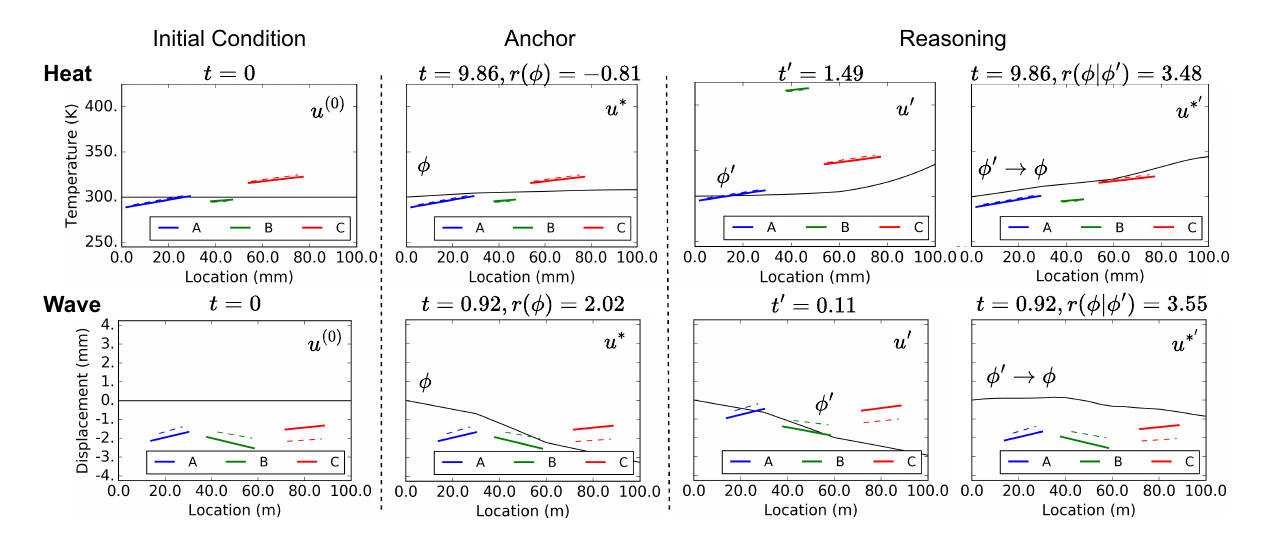}
    \vspace{-1em}
    \caption{Case study of LLM reasoning for PDE control on heat (top) and wave (bottom) problems (symbols are aligned with Fig.~\ref{fig:reasoning}).
    From left to right: 
    Directly solving $\phi$ from the initial condition $u^{(0)}$ (1st column) yields $r(\phi)$ (2nd column);
    Reasoning: solving $\phi^\prime$ from $u^{(0)}$ to get $u^\prime$ (3rd column) then solving $\phi$ from $u^\prime$ to get $r(\phi|\phi^\prime)$ (4th column).
    Black curves indicate the system's states (temperature for heat, displacement for wave) at $t$ or $t^\prime$.
    Colored segments (A, B, C) are constraints, with dashes for inequalities ($\geq$ when dashes are above the solid, and vice versa).
    Although we plot constraints, \emph{they may constrain temporal ranges} $[t_1, t_2]$ $\not\owns t, t^\prime $.
    Constraint STLs can be found in Appendix~\ref{appendix:case_study_fig6}.}
    \label{fig:case_study}
    \vspace{-0.5em}
\end{figure*}

\subsection{Improved Utility via PDE Reasoning of Controller}

Beyond autoformalization and program synthesis, our most important contribution is the scientific reasoning on PDE problems by the Controller.
In addition to the MathCoder2 and GPT models,
we consider another baseline, \emph{random sampling}, which naively generates reasoning steps by randomly sampling the anchor's constraints.

\begin{table}[t!]
\centering
\caption{Overview of our reasoning data. We threshold 3 difficulty levels of questions by the Success Rate $\overline{P}$ of random sampling.}
\vspace{-0.5em}
\resizebox{0.42\textwidth}{!}{
\begin{tabular}{lccc}
\toprule
Heat & Training & Testing & Total \\ \midrule
Num. $(\phi^{\prime(w)}, \phi^{\prime(l)})$ Pairs & 4813 & 1181 & 5994 \\
Easy $\overline{P} \in (0.8, 1)$ & 27.1\% & 26.1\% & 26.9\% \\
Medium $\overline{P} \in (0.5, 0.8]$ & 37.3\% & 37.8\% & 37.4\% \\
Hard $\overline{P} \in [0, 0.5]$ & 35.6\% & 36.2\% & 35.7\% \\ \midrule
Wave & Training & Testing & Total \\ \midrule
Num. $(\phi^{\prime(w)}, \phi^{\prime(l)})$ Pairs & 3812 & 966 & 4778 \\
Easy $\overline{P} \in (0.88, 1)$ & 32.5\% & 33.6\% & 32.7\% \\
Medium $\overline{P} \in (0.55, 0.88]$ & 33.1\% & 32.5\% & 33.0\% \\
Hard $\overline{P} \in [0, 0.55]$ & 34.4\% & 33.9\% & 34.3\% \\ \bottomrule
\end{tabular}
}
\label{table:data_reasoning}
\vspace{-0.5em}
\end{table}

\begin{table*}[t!]
\centering
\caption{
Scientific reasoning over PDE control problems via our Controller LLM. 
Deviations over 5 seeds are in parentheses.
``Valid STL $\phi^\prime$ (\%)'': ratio of valid proposed subgoal STL $\phi^\prime$ (i.e. without any syntax errors or improper time constraints).
\textbf{Bold} indicates the best, \ul{underline} denotes the runner-up. “x” indicates no valid STLs were generated for evaluation. “-” indicates not applicable.
}
\vspace{-0.5em}
\resizebox{0.99\textwidth}{!}{
\addtolength{\tabcolsep}{-0.3em}
\begin{tabular}{cc|ccccc|ccccc}
\toprule
\multirow{3}{*}{PDE} & \multirow{3}{*}{\begin{tabular}{@{}c@{}}Difficulty\\Level\end{tabular}} & \multicolumn{5}{c|}{Success Rate $\overline{P}$ ($\uparrow$)} & \multicolumn{5}{c}{Utility Gain $\overline{\Delta r}$ ($\uparrow$)} \\ \cmidrule{3-7} \cmidrule{8-12}
 &  & Ours & \begin{tabular}[c]{@{}c@{}}Random\\ Sampling\end{tabular} & \begin{tabular}[c]{@{}c@{}}Math-\\Coderv2\end{tabular} & \begin{tabular}[c]{@{}c@{}}GPT\\(o1-mini)\end{tabular} & \begin{tabular}[c]{@{}c@{}}GPT\\(4o)\end{tabular} & Ours & \begin{tabular}[c]{@{}c@{}}Random\\ Sampling\end{tabular} & \begin{tabular}[c]{@{}c@{}}Math-\\Coderv2\end{tabular} & \begin{tabular}[c]{@{}c@{}}GPT\\(o1-mini)\end{tabular} & \begin{tabular}[c]{@{}c@{}}GPT\\(4o)\end{tabular} \\ \midrule
\multirow{5}{*}{Heat} 
 & Easy & \textbf{0.966} (0.0453) & \ul{0.886} (0.0365) & 0.396 (0.2419) & x & 0.718 (0.1012) & \textbf{2.233} (0.6662) & 1.594 (1.0333) & 0.795 (07985) & x & \ul{1.614} (0.0262)  \\ \cmidrule{2-12}
 & Medium & \textbf{0.877} (0.1010) & \ul{0.694} (0.0909) & 0.340 (0.2324) & 0 (0) & 0.468 (0.0912) & \textbf{1.090} (0.5030) & \ul{0.526} (0.6986) & -0.109 (0.2659) & -0.601 (0) & 0.222 (0.0560)  \\ \cmidrule{2-12}
 & Hard & \textbf{0.592} (0.1692) & 0.356 (0.1319) & 0.236 (0.1887) & 0 (0) & \ul{0.469} (0.1147) & \textbf{1.035} (0.8486) & -0.380 (1.2688) & -0.964 (0.5101) & -1.489 (0) & \ul{0.855} (0.0611)  \\ \cmidrule{2-12}
 & All & \textbf{0.812} (0.1052) & \ul{0.645} (0.0864) & 0.324 (0.221) & x & 0.552 (0.1024) & \textbf{1.453} (0.6726) & 0.580 (1.0002) & -0.093 (0.5249) & x & \ul{0.897} (0.0477)  \\ \midrule\midrule
\multirow{5}{*}{Wave} 
 & Easy & 0.936 (0.0600) & 0.928 (0.0232) & \ul{0.954} (0.0725) & x & \textbf{1} (0) & 1.423 (0.4135) & 1.110 (0.4986) & \ul{1.601} (0.1794) & x & \textbf{1.706} (0)  \\ \cmidrule{2-12}
 & Medium & \ul{0.833} (0.1009) & 0.737 (0.0894) & 0.769 (0.0362) & x & \textbf{0.933} (0.0408) & \textbf{0.901} (0.4389) & 0.704 (0.7072) & \ul{0.830} (0.2122) & x & 0.652 (0.0191)  \\ \cmidrule{2-12}
 & Hard & \ul{0.328} (0.1220) & 0.294 (0.1620) & 0.324 (0.1036) & x & \textbf{0.386} (0.1013) & \textbf{-0.349} (0.4357) & \ul{-0.531} (0.8028) & -0.670 (0.1945) & x & -0.609 (0.0196) \\ \cmidrule{2-12}
 & All & \ul{0.699} (0.0943) & 0.653 (0.0915) & 0.682 (0.0708) & x & \textbf{0.773} (0.0474) & \textbf{0.658} (0.4293) & 0.427 (0.6695) & \ul{0.587} (0.1954) & x & 0.583 (0.0129)  \\ \midrule\midrule
 \multicolumn{2}{c|}{Valid STL $\phi^\prime$ (\%) ($\uparrow$)} & \textbf{82.70} (1.97) & - & \ul{42.45} (10.54) & 0.04 (0.10) & 2.55 (0.65) & \textbf{82.70} (1.97) & - & \ul{42.45} (10.54) & 0.04 (0.10) & 2.55 (0.65)\\
 \bottomrule
\end{tabular}
}
\label{table:controller}
\vspace{-0.5em}
\end{table*}

\paragraph{Evaluation Metrics.}

During inference, we sample from the Controller multiple times.
We evaluate the reasoning performance on PDE control problems with two metrics:
\vspace{-1em}
\begin{itemize}[leftmargin=*]
\item
Success Rate $\overline{P}$: The percentage of sampled reasoning step ($\phi^\prime$) that can improve the anchor problem ($\phi$), averaged across all anchor problems.
$\overline{P} \triangleq \mathbb{E}_{\phi} P(\phi) = \mathbb{E}_{\phi} p\left(r(\phi | \phi^\prime) > r(\phi)\right)$.
\item
Utility Gain $\overline{\Delta r}$: The expected improvement in utility via a sampled reasoning step, averaged across all anchor problems.
$\overline{\Delta r} \triangleq \mathbb{E}_{\phi}\mathbb{E}_{\phi^\prime}[r(\phi | \phi^\prime) - r(\phi)]$.
\end{itemize}

\paragraph{Difficulty Levels.}
Intuitively, some anchor problems are easy to improve via reasoning,
while others may be more challenging to improve.
To comprehensively study the performance of our controller, we design three difficulty levels based on the Success Rate, $\overline{P}$, of random sampling.
We group problems by choosing thresholds on $\overline{P}$ of random sampling such that all difficulty levels share a balanced number of problems during testing. We overview our reasoning data in Table~\ref{table:data_reasoning} and provide examples in Appendix~\ref{appendix:case_study}.

\paragraph{Results\footnotemark.}

\footnotetext[5]{In Table~\ref{table:controller}, to isolate the evaluation of the PDE reasoning quality ($\phi^\prime$) from influence by autoformalization and code generation, we provide the true Python code for any valid generated subgoal STL $\phi^\prime$.
Table~\ref{table:controller_end2end} shows end-to-end results but cannot decouple the quality of LLM PDE reasoning.
}

We observe the following from Table~\ref{table:controller}:
\vspace{-1em}
\begin{itemize}[leftmargin=*]
\item In general, our Controller consistently outperforms other models for both heat and wave problems. While second to GPT-4o by wave success rates, GPT-4o suggests very few valid STLs caused by syntax errors or time constraints that occur after the anchor problem.
\item In general, our Controller most significantly improves the utility.
For example, on heat problems, our $\overline{\Delta r} = 1.453$, which improves 62\% over the second-best (GPT 4o, $\overline{\Delta r} = 0.897$).
Although no model improves utility for ``hard''-level wave problems, our Controller still proposes the highest quality subgoals.
\item  Our Controller suggests far and away the highest ratio of valid STLs ($\phi^\prime$), almost doubling the second-ranked MathCoder2.
\item In most heat cases, MathCoder2 and GPT models are worse than random sampling. GPT o1-mini fails entirely due to invalid hallucinations in subgoal STL proposals.
\end{itemize}

\vspace{-1em}
On manual data, due to invalid subgoal STL proposals, all models fail to generate meaningful reasoning steps. Firstly, inconsistencies in the natural language of the manual data cause the models to generate invalid constraint values, such as different units to describe time within a sentence.
Secondly, new notation such as ``:='' in NL, result in unbalanced brackets, hallucinated numbers, and quantifiers.
Thirdly, the models propose invalid time constraints for reasoning steps, which should occur before the anchor's time constraints.

\paragraph{Case Study.}
To better illustrate our PDE controller, we show one case for heat and wave
in Fig.~\ref{fig:case_study}.
In both examples, first optimizing the reasoning steps
leads to new initial conditions that better solve each anchor problem.

\section{Related Works}

First, our paper belongs to the broader community of
\textbf{autoformalization in AI-for-math}.
Autoformalization converts informal math into formal logic. LLMs like GPT-3.5/4o have been used to translate problems into Isabelle/Lean4 sketches~\cite{zhou2024don,jiang2023multilingual}. VernaCopter~\cite{van2024vernacopter} autoformalizes natural languages into STL with correctness checks. Hybrid methods combine manual and automated steps~\cite{xin2024deepseek,xiong2023trigo,murphy2024autoformalizing,mishra2022lila}. We introduce LLMs trained to autoformalize informal PDE control into STL.
Second, our paper targets
\textbf{PDE Controls}.
PDEs model physical systems; control aims to ensure stability or optimize behavior~\cite{alvarez2020formal,wei2024generative}. Classical methods use adjoints~\cite{lions1971optimal}, while learning-based ones use differentiable physics~\cite{holl2020learning} or RL~\cite{rabault2019artificial}. LLMs help with multimodal PDE surrogate modeling~\cite{lorsung2025explainlikeimfive}. We reformulate PDE control into MILP via FEM~\cite{sadraddini2015robust} and use LLMs to propose initializations for open-loop control, guided by a utility score (Appendix~\ref{appendix:utility_stl}).
Finally, our work also studies \textbf{LLM-based Task Planning}.
LLMs translate natural language to formal plans (e.g., LTL)\cite{pan2023dataefficientlearningnaturallanguage}. Recent work improves plan reliability~\cite{wangconformal,ren2023robots}, with CLMASP refining outputs via ASP~\cite{lin2024clmasp}. We extend LLM-based reasoning to PDE control for the first time.

Please read Appendix~\ref{appendix:related works} for more related works.

\section{Conclusion}

In this paper, we aim to transform the automation of PDE control problems.
By utilizing LLMs to interpret natural language problem descriptions, formalize mathematical expressions, and apply scientific reasoning, we demonstrate that LLMs can fully automate PDE control while even improving control performance.
Our research emphasizes the importance of LLMs in applied mathematics, and seeks to enable more accessible, scalable, and robust PDE solutions, ultimately expanding the practical reach and reliability of PDE applications across scientific and engineering domains.

\section*{Acknowledgements}
We thank Dr. Danqi Chen for her helpful comments.
We thank all participants in our questionnaire for manually writing PDE control problems. For privacy reasons, we do not disclose their names.

\section*{Impact Statement}

This paper presents work whose goal is to advance the field of 
Machine Learning. There are many potential societal consequences 
of our work, none of which we feel must be specifically highlighted here.

\bibliography{example_paper}
\bibliographystyle{icml2025}

\newpage
\appendix
\onecolumn

\section{More Background on Formal Methods for PDE Control}
\label{appendix:pde_control_background}

\subsection{PDEs}
\label{appendix:pdes}
    We consider controlling systems governed by two popular PDEs (in 1D space):
    \begin{itemize}[leftmargin=*]
        \setlength\itemsep{0em}
        \item Heat Equation: Describes how heat diffuses through a material over time. Applications: temperature in buildings, pollution in the environment.
\begin{equation}
\begin{aligned}
& \rho c \frac{\partial \bm{u}}{\partial t}-\kappa \frac{\partial^2 \bm{u}}{\partial x^2}=0, \\
& \kappa \frac{\partial \bm{u}}{\partial x}(L, t)=\bm{q}(t), \\
& \bm{u}(0, t) = g_0, \qquad \qquad\forall t \in[0, t_{\max}], \\
& \bm{u}(x, 0) = u_0(x), \qquad \,\, \forall x \in [0, L] .
\end{aligned}
\end{equation}
$\rho, c, \kappa>0$: density, specific thermal capacity, thermal conductivity of the material respectively.
$\bm{u}$ the spatiotemporal temperature of the material.
$x \in [0, L]$ is the spatial location.
$t \in [0, t_{\max}]$ is the time.
$\bm{q}$ is the time-variant external heat source, applied at $x=L$.
$g_0$ is the boundary condition applied at $x=0$.
$u_0$ is the initial condition (temperature).

        \item Wave Equation: Models the propagation of waves (sound, electromagnetic, or water waves). Applications: Acoustic control, vibration control in structures (e.g., bridges, buildings).
\begin{equation}
\begin{aligned}
& \rho \frac{\partial^2 \bm{u}}{\partial t^2}-E \frac{\partial^2 \bm{u}}{\partial x^2}=0, \\
& E \frac{\partial \bm{u}}{\partial x}(L, t) = \bm{F}(t), \\
& \bm{u}(0, t) = g_0, \qquad \qquad \forall t \in [0, t_{\max}], \\
& \bm{u}(x, 0) = u_0(x), \qquad \,\, \forall x \in [0, L] .
\end{aligned}
\end{equation}
$\rho, E>0$: density, Young's Modulus of the material respectively.
$\bm{u}$ the spatiotemporal displacement of the material.
$x \in [0, L]$ is the spatial location.
$t$ is the time.
$\bm{F}$ is the time-variant external fource, applied at $x=L$.
$g_0$ is the boundary condition applied at $x=0$.
$u_0$ is the initial condition (displacement), and typically we set it as 0.
    \end{itemize}

\subsection{Utility of STL}
\label{appendix:utility_stl}

The continuous utility value of STL $r(\phi)$ is calculated with the following cases and rules:
\begin{align}
& r\left(\bm{u}, \bm{u}\geq ax+b, t\right) = \bm{u}(x, t) - (ax+b) \label{eq:stl_utility_1} \\
& r\left(\bm{u}, \bm{u}\leq ax+b, t\right) = (ax+b) - \bm{u}(x, t) \label{eq:stl_utility_2} \\
& r\left(\bm{u}, \phi_{1} \wedge \phi_{2}, t\right) = \min \left\{r\left(\bm{u}, \phi_{1}, t\right), r\left(\bm{u}, \phi_{2}, t\right)\right\} \\
& r\left(\bm{u}, \phi_{1} \vee \phi_{2}, t\right) = \max \left\{r\left(\bm{u}, \phi_{1}, t\right), r\left(\bm{u}, \phi_{2}, t\right)\right\} \\
& r\left(\bm{u}, \mathbf{F}_{[a, b)} \phi, t\right) = \sup _{t_{f} \in[t+a, t+b)}\left\{r\left(\bm{u}, \phi, t_{f}\right)\right\} \\
& r\left(\bm{u}, \mathbf{G}_{[a, b)} \phi, t\right) = \inf _{t_{g} \in[t+a, t+b)}\left\{r\left(\bm{u}, \phi, t_{g}\right)\right\}
\end{align}
Here, Eq.~\ref{eq:stl_utility_1} and~\ref{eq:stl_utility_2} indicates the linear constraint we consider in Eq.~\ref{eq:stl}.

\subsection{MILP Formulation of Control Synthesis}
\label{appendix:milp_formulation}

Solving the PDE control problem can be relaxed and formulated into a PDE-constrained optimization problem, which can be further solved by mixed-integer linear programming (MILP).
We brief the high-level steps, and we recommend readers to read~\cite{sadraddini2015robust,alvarez2020formal} for more details.

We start with spatially and temporally discretizing the PDE.
This is achieved by the finite element method (FEM):
\begin{enumerate}[leftmargin=*]
\item
The PDE is converted into its weak (variational) form by integrating against suitable test functions $\bm{v}(x)$.
The purpose is to reduce the second-order derivatives of $\bm{u}$ to first derivatives so that we can obtain (linear) approximations to $\bm{u}$.
This also simplifies boundary condition handling and smoothness requirements in the original PDE.

\item
The spatial domain of the PDE's weak form is discretized by dividing it into small, simple geometric elements (intervals in 1D, triangles/quadrilaterals in 2D; tetrahedra in 3D), essentially forming a mesh, and choosing local basis functions on each element.

\item
Contributions from local (spatial) elements are assembled across the entire mesh into a global linear system, often written as $\bm{M} \dot{\tilde{\bm{u}}} + K \tilde{\bm{u}} = \bm{F}$.
The stiffness matrix ($\bm{K}$) and mass matrix ($\bm{M}$) encode the PDE's structure.
The banded stiffness matrix ($\bm{K}$) arises from terms involving derivatives (e.g., $\nabla \bm{u} \cdot \nabla \bm{v}$ in the weak form).
The diagonal mass matrix ($\bm{M}$) comes from non-derivative terms (e.g., $\bm{u} \cdot \bm{v}$).
$F$ is the ``force'' vector (from external source terms/boundary conditions like heat and force).
This step transforms the PDE into an ordinary differential equation (ODE) in time for the discretized spatial domain.

\item
The temporal domain is further discretized using finite difference schemes, obtaining a set of difference equations that must be solved at each time step.
This final step produces the final linear (or nonlinear) system of equations that is solved numerically to approximate the solution of the original PDE.
\begin{equation}
\bm{M} \frac{\tilde{\bm{u}}^{n+1}-\tilde{\bm{u}}^n}{\Delta t} + \bm{K} \tilde{\bm{u}}^{n+1} = \bm{F}^{n+1}.
\end{equation}
This can be further rearranged to a linear system:
\begin{equation}
(\bm{M} + \Delta t \bm{K}) \tilde{\bm{u}}^{k+1} = \bm{M} \tilde{\bm{u}}^k + \Delta t \bm{F}^{k+1}.
\end{equation}
\end{enumerate}

At this moment, we can re-formulate the original PDE control problem into the following PDE-constrained optimization problem:
\begin{align}
& \max r\left(\phi, \tilde{\bm{u}}\right) \label{eq:milp_object} \\
\text{s.t. } &
(\bm{M} + \Delta t \bm{K}) \tilde{\bm{u}}^{k+1} = \bm{M} \tilde{\bm{u}}^k + \Delta t \bm{F}^{k+1}, \label{eq:milp_constraint} \\
& \tilde{\bm{u}}^{0} = \tilde{\bm{u}}(0). \notag
\end{align}
This formulation is equivalent to an MILP problem because:
\begin{itemize}[leftmargin=*]
\item $\phi$ is only applied to limited spatiotemporal ranges. After the discretization of PDE, essentially $\phi$ is only selectively applied to certain areas of grids over our (1D) mesh. That means, we need \textbf{binary} variables to encode the absence/presence of $\phi$ over the spatiotemporal domain.
\item 
The discretized PDE constraints Eq.~\ref{eq:milp_constraint} is linear in $\tilde{\bm{u}}$.
Additionally, based on Appendix~\ref{appendix:utility_stl}, the objective function $r(\phi, \tilde{\bm{u}})$ (Eq.~\ref{eq:milp_object}) is also linear in $\phi$ and $\tilde{\bm{u}}$.
\end{itemize}

This MILP problem is non-convex, due to
the $\min, \max$ operation and non-differentiability (Appendix~\ref{appendix:utility_stl}) of the objective function $r\left(\phi, \tilde{\bm{u}}\right)$ in Eq.~\ref{eq:milp_object}.
This MILP problem is solved using the off-the-shelf Gurobi solver~\cite{gurobi}.

\subsection{Metrics - Additional Design Considerations}

The careful design of our utility score, in Appendix \ref{appendix:utility_stl}, can faithfully quantify whether the STLs (constraints) are fully met by the solution simulated by the Gurobi solver.
This utility score is inherited from prior work~\cite{alvarez2020formal}.

Compared with standard metrics such as pass@k, our metrics can better quantify our LLMs’ performance. IoU is designed to provide fine-grained quantification of the logic structures (junctions, temporal operators, boundary constraints) while pass@k lacking nuance will fail or succeed from trivial text-level differences in STL. 
We further explain the connection between our metrics and pass@k below:
\begin{itemize}
    \item For autoformalization (Translator), IoU is equivalent to “average@k”, it averages the alignment between predictions and targets over multiple generations and problems. We can discretize IoU into “pass@k” by considering only pass/fail cases based on token-level differences; but this ignores fine-grained quantifications of autoformalization. Namely, in tables \ref{table:heat-pass-at-k} and \ref{table:wave-pass-at-k} below, IoU and pass@k are not always aligned.
    \item For code generation (Coder), the executability metric is essentially pass@k from the executability perspective.
\end{itemize}

\begin{table}[h!]
\centering
\caption{Performance comparison for autoformalization (Translator) on Heat. Deviations over 3 seeds are in parentheses.}\label{table:heat-pass-at-k}
\resizebox{0.57\textwidth}{!}{
\addtolength{\tabcolsep}{-0.3em}
\begin{tabular}{ccccc}
\toprule
Model & IoU & Pass@1 & Pass@2 & Pass@3 \\ 
\midrule
Ours & 0.992 (0.07) & 0.978 (0.142) & 0.980 (0.134) & 0.982 (0.131) \\ 
MathCoder2 & 0.772 (0.35) & 0.538 (0.480) & 0.565 (0.484) & 0.583 (0.493) \\ 
\bottomrule
\end{tabular}
}
\label{table:heat_performance}
\end{table}

\begin{table}[h!]
\centering
\caption{Performance comparison for autoformalization (Translator) on Wave. Deviations over 3 seeds are in parentheses.}\label{table:wave-pass-at-k}
\resizebox{0.58\textwidth}{!}{
\addtolength{\tabcolsep}{-0.3em}
\begin{tabular}{ccccc}
\toprule
Model & IoU & Pass@1 & Pass@2 & Pass@3 \\ 
\midrule
Ours & 0.992 (0.07) & 0.971 (0.161) & 0.975 (0.152) & 0.977 (0.149) \\ 
MathCoder2 & 0.772 (0.35) & 0.3305 (0.4396) & 0.3726 (0.4663) & 0.3971 (0.4893) \\ 
\bottomrule
\end{tabular}
}
\label{table:wave_performance}
\end{table}

\section{Case Examples and Visualizations}
\label{appendix:case_study}

\begin{figure*}[t!]
    \centering
    \includegraphics[width=1.\linewidth]{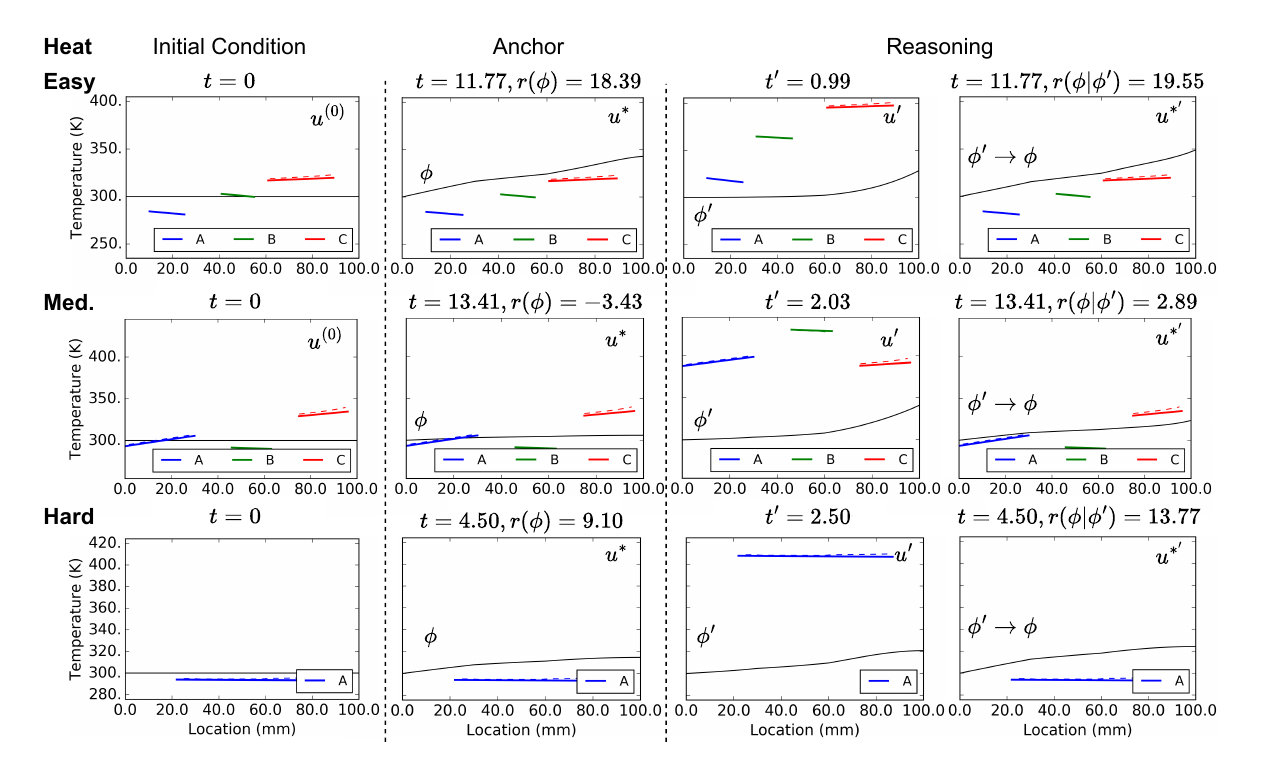}
    
    \caption{Case study of \textbf{heat} problems with different difficulty levels: easy (top), medium (middle), hard (bottom).
    Symbols are aligned with Fig.~\ref{fig:reasoning}.
    From left to right: 
    Directly solving $\phi$ from the initial condition $u^{(0)}$ (1st column) yields $r(\phi)$ (2nd column).
    Reasoning: solving $\phi^\prime$ from $u^{(0)}$ to get $u^\prime$ (3rd column) then solving $\phi$ from $u^\prime$ to get $r(\phi|\phi^\prime)$ (4th column)
    Black curves indicate the system's states (temperature for heat, displacement for wave) at $t$ or $t^\prime$.
    Colored segments are constraints, with dashes for inequalities ($\geq$ when dashes are above the solid, $\leq$ when dashes are below the solid). Note that although we always plot constraints (segments), \emph{they actually constrain different temporal ranges} $[t_1, t_2]$ and it is possible that $t, t^\prime \notin [t_1, t_2]$.}
    \label{fig:case_study_heat_nc123}
\end{figure*}

To better illustrate our PDE reasoning, we show more cases with visualizations and their corresponding STLs (constraints). All time constraints are rounded to two decimal places, and parameters that describe the linear profiles are rounded to four decimal places.

\subsection{Heat}

We show easy/medium/hard problems in Fig.~\ref{fig:case_study_heat_nc123}, with their anchor STL ($\phi$) and subgoal STL ($\phi^\prime$) listed below.

\textbf{1) Easy:}

Anchor Constraints STL ($\phi$):
$$
\begin{aligned}
    &F_{[1.17, 3.48]} (\forall x \in [10, 25]  (u(x) - (-0.2169 \cdot x + 286.5171) > 0)) \land \\
    &  (G_{[4.64, 5.13]} (\forall x \in [41, 55]  (u(x) - (-0.2225 \cdot x + 311.8826) < 0)) \lor \\ 
    &   F_{[6.04, 11.77]} (\forall x \in [61, 89]  (u(x) - (0.0988 \cdot x + 310.7904) > 0)))
\end{aligned}
$$  

Subgoal STL Proposal ($\phi^\prime$) by Controller:

$$
\begin{aligned}
    &F_{[0.42, 0.99]} (\forall x \in [10, 25]  (u(x) - (-0.2907 \cdot x + 323.3970) > 0)) \land \\
    &  (G_{[0.10, 0.57]} (\forall x \in [31, 46]  (u(x) - (-0.1338 \cdot x + 368.9958) < 0)) \lor \\ 
    &   F_{[0.30, 0.96]} (\forall x \in [61, 89]  (u(x) - (0.0788 \cdot x + 390.7948) > 0)))
\end{aligned}
$$

\textbf{2) Medium:}

Anchor Constraints STL ($\phi$):
$$
\begin{aligned}
    &G_{[2.18, 2.70]} (\forall x \in [0, 30]  (u(x) - (0.4159 \cdot x + 293.2549) > 0)) \lor \\ 
    &(G_{[4.03, 7.79]} (\forall x \in [46, 63]  (u(x) - (-0.0956 \cdot x + 296.0596) < 0)) \land \\
    &F_{[8.33, 13.41]} (\forall x \in [75, 96]  (u(x) - (0.2602 \cdot x + 309.7111) > 0)))
\end{aligned}
$$

Subgoal STL Proposal ($\phi^\prime$) by Controller:
$$
\begin{aligned}
    &G_{[0.66, 1.68]} (\forall x \in [0, 30]  (u(x) - (0.3616 \cdot x + 387.4454) > 0)) \lor \\ 
    &(G_{[0.52, 2.03]} (\forall x \in [46, 63]  (u(x) - (-0.1061 \cdot x + 435.4267) < 0)) \land \\
    &F_{[0.22, 1.17]} (\forall x \in [75, 96]  (u(x) - (0.1779 \cdot x + 374.4556) > 0)))
\end{aligned}
$$

\textbf{3) Hard:}

Anchor Constraints STL ($\phi$):

$$
\begin{aligned}
    G_{[2.62, 4.50]} (\forall x \in [22, 87] (u(x) - (-0.0122 \cdot x + 294.2976) > 0))
\end{aligned}$$

Subgoal STL Proposal ($\phi^\prime$) by Controller:
$$
\begin{aligned}
    G_{[1.29, 2.50]} (\forall x \in [22, 87]  (u(x) - (-0.0157 \cdot x + 408.1535) > 0))
\end{aligned}
$$

\begin{figure*}[t!]
    \centering
    \includegraphics[width=0.98\linewidth]{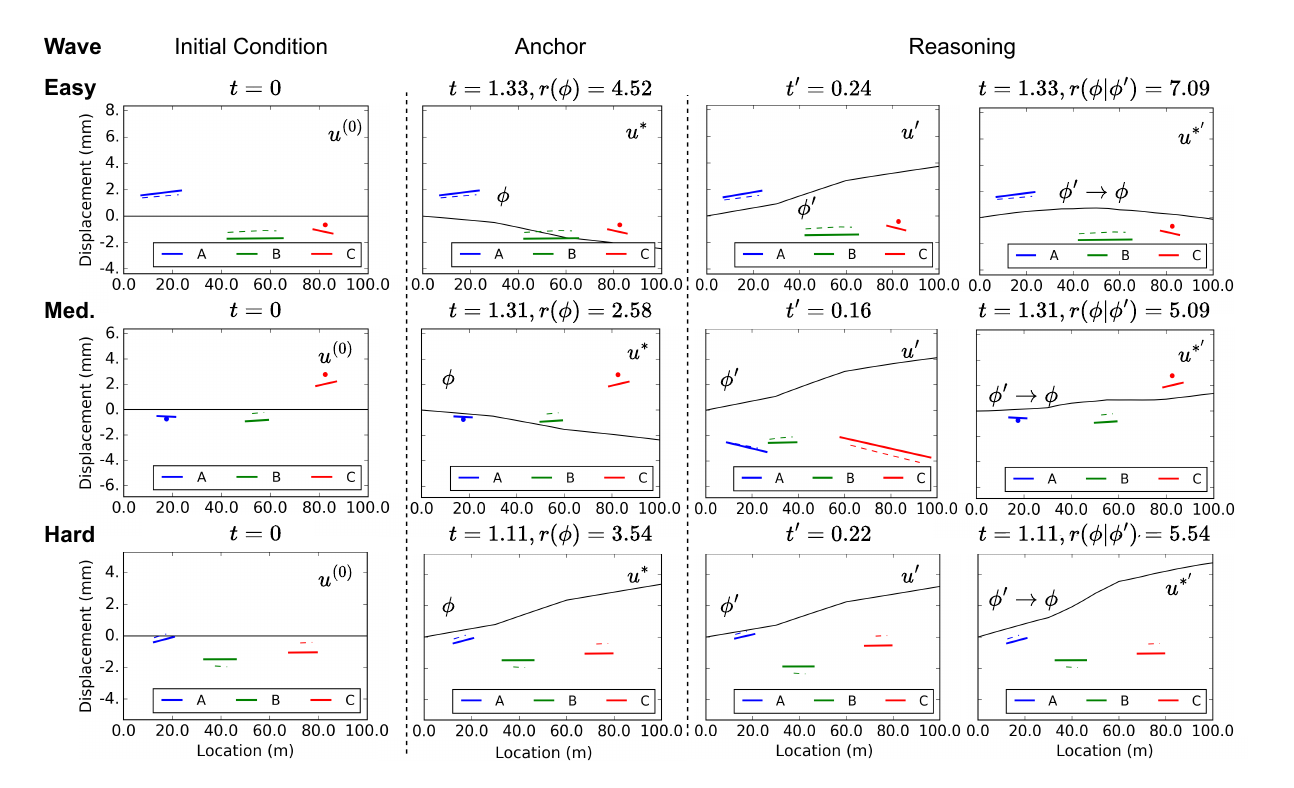}
    \caption{Case study of \textbf{wave} problems with different difficulty levels: easy (top), medium (middle), hard (bottom).
    Symbols are aligned with Fig.~\ref{fig:reasoning}.
    From left to right: 
    Directly solving $\phi$ from the initial condition $u^{(0)}$ (1st column) yields $r(\phi)$ (2nd column).
    Reasoning: solving $\phi^\prime$ from $u^{(0)}$ to get $u^\prime$ (3rd column) then solving $\phi$ from $u^\prime$ to get $r(\phi|\phi^\prime)$ (4th column)
    Black curves indicate the system's states (temperature for heat, displacement for wave) at $t$ or $t^\prime$.
    Colored segments are constraints, with dashes for inequalities ($\geq$ when dashes are above the solid, $\leq$ when dashes are below the solid). Note that although we always plot constraints (segments), \emph{they actually constrain different temporal ranges} $[t_1, t_2]$ and it is possible that $t, t^\prime \notin [t_1, t_2]$.}    
    \label{fig:case_study_wave_nc123}
\end{figure*}

\subsection{Wave}

We show easy/medium/hard problems in Fig.~\ref{fig:case_study_wave_nc123}, with their anchor STL ($\phi$) and subgoal STL ($\phi^\prime$) listed below.

\textbf{1) Easy:}

Anchor Constraints STL ($\phi$):
$$
\begin{aligned}
    &(G_{[0.25, 0.54]} (\forall x \in [7207, 23479]  (u(x) - (2.2684e-05 \cdot x + 1.4129) < 0)) \land \\
    &F_{[0.76, 0.84]} (\forall x \in [42469, 65095]  (u(x) - (1.8952e-06 \cdot x - 1.7928) > 0))) \lor \\ 
    &F_{[1.12, 1.33]} (\forall x \in [77653, 85444]  (u(x) - (-4.0675e-05 \cdot x + 2.1560) > 0))
\end{aligned}
$$

Subgoal STL Proposal ($\phi^\prime$) by Controller:
$$
\begin{aligned}
    &(G_{[0.10, 0.24]} (\forall x \in [7207, 23479]  (u(x) - (3.0242e-05 \cdot x + 1.1961) < 0)) \land \\
    &F_{[0.05, 0.09]} (\forall x \in [42469, 65095]  (u(x) - (2.3575e-06 \cdot x - 1.5473) > 0))) \lor \\ 
    &F_{[0.10, 0.18]} (\forall x \in [77653, 85444]  (u(x) - (-4.4208e-05 \cdot x + 2.7018) > 0))
\end{aligned}
$$

\textbf{2) Medium:}

Anchor Constraints STL ($\phi$):
$$
\begin{aligned}
    &G_{[0.10, 0.24]} (\forall x \in [13787, 21080]  (u(x) - (-9.5400e-06 \cdot x - 0.3744) < 0)) \lor \\
    &F_{[0.05, 0.09]} (\forall x \in [49923, 59039]  (u(x) - (1.2003e-05 \cdot x -1.5231) > 0)) \lor \\ 
    &G_{[0.78, 1.31]} (\forall x \in [78762, 86964]  (u(x) - (4.3983e-05 \cdot x -1.5994) > 0))
\end{aligned}
$$

Subgoal STL Proposal ($\phi^\prime$) by Controller:

$$
\begin{aligned}
    &G_{[0.06, 0.16]} (\forall x \in [9084, 26246]  (u(x) - (-4.3491e-05 \cdot x - 2.1348) > 0)) \land \\
    &G_{[0.04, 0.04]} (\forall x \in [27204, 39168]  (u(x) - (4.2688e-06 \cdot x - 2.6843) > 0)) \land \\
    &G_{[0.01, 0.07]} (\forall x \in [58194, 97070]  (u(x) - (-4.0965e-05 \cdot x + 0.2641) < 0))
\end{aligned}
$$

\textbf{3) Hard:}

Anchor Constraints STL ($\phi$):
$$
\begin{aligned}
    &(G_{[0.23, 0.30]} (\forall x \in [12400, 20684]  (u(x) - (4.0369e-05 \cdot x - 0.9002) > 0)) \land \\
    &F_{[0.72, 0.82]} (\forall x \in [33059, 46052]  (u(x) - (3.5491e-07 \cdot x -1.4933) < 0))) \lor \\ 
    &F_{[1.10, 1.11]} (\forall x \in [67963, 79313]  (u(x) - (1.6090e-06 \cdot x -1.1675) > 0))
\end{aligned}
$$

Subgoal STL Proposal ($\phi^\prime$) by Controller:
$$
\begin{aligned}
    &(G_{[0.07, 0.22]} (\forall x \in [12400, 20684]  (u(x) - (3.4256e-05 \cdot x -0.5172) > 0)) \land \\
    &F_{[0.04, 0.11]} (\forall x \in [33059, 46052]  (u(x) - (2.9472e-07 \cdot x -1.8896) < 0))) \lor \\ 
    &F_{[0.01, 0.20]} (\forall x \in [67963, 79313]  (u(x) - (2.3060e-06 \cdot x -0.7148) > 0))
\end{aligned}
$$

\subsection{STLs for Examples in Fig.~\ref{fig:case_study}}
\label{appendix:case_study_fig6}
\textbf{Heat:}

Anchor Constraints STL ($\phi$) shown as A, B, C respectively in the \textit{Initial Condition, Anchor} and right \textit{Reasoning} columns of Fig.~\ref{fig:case_study}:
$$
\begin{aligned}
&G_{[1.63, 3.13]} (\forall x \in [2, 29]  (u(x) - (0.4565 \cdot x + 287.7909) > 0)) \lor \\  
&(G_{[4.25, 4.58]} (\forall x \in [38, 47]  (u(x) - (0.2137 \cdot x + 287.1038) < 0)) \land \\ 
&F_{[5.94, 9.86]} (\forall x \in [54, 77]  (u(x) - (0.3008 \cdot x + 299.3877) > 0)))
\end{aligned}
$$

Subgoal STL Proposal ($\phi^\prime$) by Controller shown as A, B, C respectively in the left \textit{Reasoning} column of Fig.~\ref{fig:case_study}:
$$
\begin{aligned}
&G_{[0.69, 1.49]} (\forall x \in [2, 29]  (u(x) - (0.4088 \cdot x + 294.5123) > 0)) \lor \\  
&(G_{[0.26, 0.33]} (\forall x \in [38, 47]  (u(x) - (0.2907 \cdot x + 404.7615) < 0)) \land \\ 
&F_{[0.06, 0.10]} (\forall x \in [54, 77]  (u(x) - (0.3503 \cdot x + 316.1354) > 0)))
\end{aligned}
$$

\textbf{Wave:}

Anchor Constraints STL ($\phi$) shown as A, B, C respectively in the \textit{Initial Condition, Anchor} and right \textit{Reasoning} columns of Fig.~\ref{fig:case_study}:
$$
\begin{aligned}
& (G_{[0.16, 0.20]} (\forall x \in [14057, 29980]  (u(x) - (2.8994e-05 \cdot x - 2.5372) > 0)) \land \\
& G_{[0.28, 0.37]} (\forall x \in [38096, 58208]  (u(x) - (-2.9597e-05 \cdot x - 0.8070) > 0))) \lor \\ 
& F_{[0.45, 0.92]} (\forall x \in [71793, 88339]  (u(x) - (1.2523e-05 \cdot x - 2.4337) < 0))
\end{aligned}
$$

Subgoal STL Proposal ($\phi^\prime$) by Controller shown as A, B, C respectively in the left \textit{Reasoning} column of Fig.~\ref{fig:case_study}:
$$
\begin{aligned}
& (G_{[0.00, 0.01]} (\forall x \in [14057, 29980]  (u(x) - (3.0385e-05 \cdot x - 1.3785) > 0)) \land \\
& G_{[0.03, 0.07]} (\forall x \in [38096, 58208]  (u(x) - (-2.2655e-05 \cdot x - 0.5356) > 0))) \lor \\ 
& F_{[0.00, 0.11]} (\forall x \in [71793, 88339]  (u(x) - (1.6430e-05 \cdot x - 1.7368) < 0))
\end{aligned}
$$

\subsection{Solutions to Examples in Fig.~\ref{fig:case_study}}
\begin{figure*}[t!]
    \centering
    \includegraphics[width=0.8\linewidth]{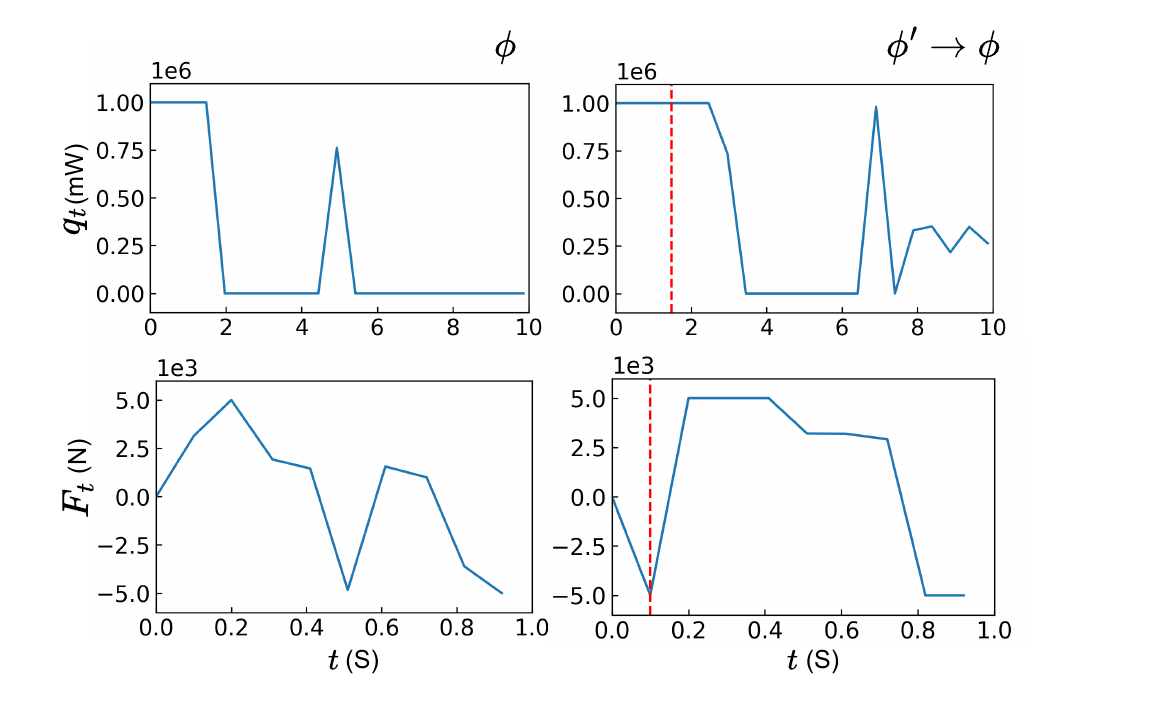}
    \caption{Synthesized control inputs for the \textbf{heat} (top) and the \textbf{wave} (bottom) problems in Fig.\ref{fig:case_study}.
    Left: Solution for directly solving $\phi$. 
    Right: Solution for solving $\phi$ based on subgoal $\phi'$,
    where \textcolor{red}{red} vertical dashes indicate the control shift from $\phi^\prime$ to $\phi$.
    }
    \label{fig:pde_solution}
\end{figure*}

Fig.\ref{fig:pde_solution} presents the optimized control inputs for the \textbf{heat} and \textbf{wave} problems shown in Fig.\ref{fig:case_study}. The figure illustrates how the heat source ($q_t$) and the force ($F_t$) applied at the end of the rod change over time. We provide the solutions for 2 cases, the first column shows the solution obtained by directly solving the problem $\phi$, while the second column shows the solution to solve the problem $\phi$ after the subgoal $\phi'$.

\section{Dataset Details}

\subsection{PDE Parameter Ranges}
\label{appendix:parameter_ranges}

For generating our dataset, we sample hyperparameters (that define control problems) in the following range (Table~\ref{table:parameters_heat} and~\ref{table:parameters_wave}).
Note that our heat and wave problems are all 1D, making some dimension-related units different from those in 3D.

\begin{table}[h!]
\centering
\caption{Ranges for hyperparameters used in heat problems.}
\vspace{-0.5em}
\resizebox{0.99\textwidth}{!}{
\addtolength{\tabcolsep}{-0.3em}
\begin{tabular}{|c|c|c|c|c|c|c|c|c|c|c|}
    \hline
    \begin{tabular}{@{}c@{}} Rod Length \\ ($mm$) \end{tabular} & \begin{tabular}{@{}c@{}} Fixed Temp. \\ ($K$) \end{tabular}& \begin{tabular}{@{}c@{}} Max Time \\ ($s$) \end{tabular}  & \multicolumn{2}{c|}{Linear Profile Param.} & \multicolumn{2}{c|}{\begin{tabular}{@{}c@{}}Thermal Conductivity \\($mW \cdot mm/ K$)\end{tabular}} & \multicolumn{2}{c|}{\begin{tabular}{@{}c@{}} Density \\($kg/mm$)\end{tabular}} & \multicolumn{2}{c|}{\begin{tabular}{@{}c@{}}Specific Heat Capacity \\ ($\mu J/kg/K$)\end{tabular}}\\
    \hline
     $L$ & $temp$& $t_{\max}$ & $a$ & $b$ & $\kappa_a$ ($\times 10^6$) & $\kappa_b$ ($\times 10^6$) & $\rho_a$ ($\times 10^{-6}$) & $\rho_b$ ($\times 10^{-6}$) & $c_a$ ($\times 10^8$) & $c_b$ ($\times 10^8$) \\ 
    \hline
    [50, 300] & [250, 350] & [5, 15] & [-0.5, 0.5] & $temp$ + [-20, 20] & [1.2, 1.8] & [0.4, 1.2] & [3, 6] & [3, 6] & [3, 4.5] & [4.5, 4.8] \\ 
    \hline
\end{tabular}
}
\label{table:parameters_heat}
\vspace{-0.5em}
\end{table}

\begin{table}[h!]
\centering
\caption{Ranges for hyperparameters used in wave problems.}
\vspace{-0.5em}
\resizebox{0.77\textwidth}{!}{
\addtolength{\tabcolsep}{-0.3em}
\begin{tabular}{|c|c|c|c|c|c|c|c|}
    \hline
    \begin{tabular}{@{}c@{}} Rod Length \\ ($mm$) \end{tabular} & \multicolumn{2}{c|}{\begin{tabular}{@{}c@{}} Density \\($kg/mm$)\end{tabular}} & \multicolumn{2}{c|}{\begin{tabular}{@{}c@{}} Young's Modulus \\ ($N$) \end{tabular}} & \begin{tabular}{@{}c@{}} Max Time \\ ($s$) \end{tabular} & \multicolumn{2}{c|}{Linear Profile Param.} \\
    \hline
    $L$ & $\rho_{\text{steel}}$ ($\times 10^{-6}$) & $\rho_{\text{brass}}$ ($\times 10^{-6}$) & $E_{\text{steel}}$ ($\times 10^8$) & $E_{\text{brass}}$ ($\times 10^8$) & $t_{\max}$ & $a$ ($\times 10^{-5}$) & $b$ \\ 
    \hline
    [60000, 140000] & [7.6, 8] & [8.4, 8.8] & [2, 2.4] & [1, 1.8] & [0.5, 2] & [-5, 5] & [-3, 3] \\ 
    \hline
\end{tabular}
}
\label{table:parameters_wave}
\vspace{-0.5em}
\end{table}

\subsection{Rules for Data Generation}
\label{appendix:data_generation_rules}

\paragraph{From STL to Natural Language.}
Each natural language problem consists of two parts: one part defines the premises, such as the material density, initial temperature, and rod length; the other part describes the constraints, which can be expressed as an STL formula.

To convert the STL formula into informal language, each constraint and condition is mapped to corresponding phrases. For instance, the constraint conditions $\mathbf{F}$ and $\mathbf{G}$ are described as “for one point during” and “for all time between”,  respectively. The comparison conditions are mapped based on the problem type: for heat problems, $>$ indicates “the temperature distribution of the rod should be \textit{greater than}”; for wave problems, $>$ indicates “the displacement of the rod should be \textit{stretched over}”.

We then consider the $\lor$ and $\land$ logical operators after converting each individual constraint.
For problems with two constraints, we introduce transition words such as “moreover” for $\land$ and “either…or…” for $\lor$.
For problems with three constraints, we design templates that account for the hierarchy of constraints based on the placement of parentheses in the STL formula that defines the logical order.

For example, given the STL syntax $(\mathbf{A} \land \mathbf{B}) \lor \mathbf{C}$, the template is: “Either satisfy the conditions that \textbf{A} and also \textbf{B}; or satisfy the condition that \textbf{C}.”
For $\mathbf{A} \land (\mathbf{B} \lor \mathbf{C})$, the template is: “Satisfy \textbf{A}. Afterwards, either consider \textbf{B} or \textbf{C}.”

\paragraph{From STL to Python.}

To parse the predicted STL into Python in Table~\ref{table:controller}, we first extract logical connectives ($\lor, \land$), rod intersections, and constraint equations. 
Based on the number of constraints, the corresponding variables are inserted into a Python script template. The time intervals and constraint conditions ($\mathbf{G}, \mathbf{F}$) are passed directly into the script to preserve the original STL syntax.
The generated outputs may fail to convert due to hallucinations or syntax errors.

For example, given the \textbf{STL Logic} below:
$$
\begin{aligned}
&\mathbf{G}_{[0.049, 0.053]} (\forall x \in [9829, 19907]  (u(x) - (1.882e-05 \cdot x + 0.187) < 0)) \land \\
&\mathbf{F}_{[0.051, 0.149]} (\forall x \in [40199, 56082]  (u(x) - (3.356e-06 \cdot x + -0.510) < 0))) \lor \\
&\mathbf{F}_{[0.061, 0.169]} (\forall x \in [75646, 98769]  (u(x) - (-1.390e-05 \cdot x + 2.844) > 0)
\end{aligned}
$$
We convert into its \textbf{Python snippet}:
\begin{lstlisting}
    apc0 = logic.APCont([9829, 19907], "<", 
            lambda x: 1.882e-05 * x + 0.187, lambda x: 1.882e-05)
    apc1 = logic.APCont([40199, 56082], "<", 
            lambda x: 3.356e-06 * x + -0.510, lambda x: 3.356e-06)
    apc2 = logic.APCont([75646, 98769], ">", 
            lambda x: -1.390e-05 * x + 2.844, lambda x: -1.390e-05)
    cregions = {"A": apc0,"B": apc1,"C": apc2}
    cspec = "(((G_[0.049, 0.053] (A)) &
               (F_[0.051, 0.149] (B))) |
               (F_[0.061, 0.169] (C)))"
\end{lstlisting}

\subsection{ChatGPT Augmentation}
\label{appendix:chatgpt_augmentation}
We adopt the data augmentation pipeline proposed by \citep{pan2023dataefficientlearningnaturallanguage}, where ChatGPT 4o-mini is prompted to rephrase each natural language problem five times. The problems are formulated based on STL logic with PDE parameter variables enclosed in squared brackets, which are further replaced with randomly values sampled from ranges in Appendix~\ref{appendix:parameter_ranges}. 
For example:

\textbf{Prompt}: Rephrase the source sentence in 5 different ways. Make the outputs as diverse as possible. Make sure not to change the equations starting with mu and parameters in the squared brackets, and maintain all the parameters while rephrasing.

\textbf{Source (original problem)}:
For one point during the time interval [t0] and [t1], the temperature distribution of the rod should be larger than the linear profile mu0(x) = [a0] * x + [b0] between section [r0] and [r1].

\textbf{GPT Outputs}:

1. At a specific moment within the time period [t0] to [t1], the temperature distribution along the rod must exceed the linear profile mu0(x) = [a0] * x + [b0] in the segment defined by [r0] and [r1].

2. Within the time frame of [t0] to [t1], there should be at least one instance where the rod's temperature distribution surpasses the linear profile mu0(x) = [a0] * x + [b0] from [r0] to [r1].

3. During the interval between [t0] and [t1], there is a moment at which the temperature across the rod must be greater than the linear profile mu0(x) = [a0] * x + [b0] within the range of [r0] and [r1].

4. Throughout the duration from [t0] to [t1], there exists a point in time where the temperature distribution of the rod should be greater than the linear profile mu0(x) = [a0] * x + [b0] applicable between the sections [r0] and [r1].

5. At some time during the interval between [t0] and [t1], it is required that the temperature distribution of the rod exceeds the linear profile mu0(x) = [a0] * x + [b0] in the region from [r0] to [r1].

Table~\ref{table:edit_distance_gpt} shows the character-level relative edit distance between each natural language problem and its 5 augmentations based on the problem type and the number of constraints.

\begin{table}[h!]
\centering
\caption{Averaged relative edit distance between the problem's natural language and 5 GPT augmentions. Deviations are in parentheses.}
\vspace{-0.5em}
\resizebox{0.48\textwidth}{!}{
\begin{tabular}{c|ccc}
\toprule
\multirow{4}{*}{PDE} & \multicolumn{3}{c}{Relative Edit Distance} \\ \cmidrule{2-4} & \multicolumn{3}{c}{Number of Constraints}\\
   & 1 & 2 & 3 \\ \midrule
Heat & 0.431 (0.0545) & 0.456 (0.0505) & 0.490 (0.0398) \\  
\midrule
Wave & 0.455 (0.0524) & 0.472 (0.0433) & 0.484 (0.0352) \\ 
\bottomrule
\end{tabular}
}
\label{table:edit_distance_gpt}
\end{table}

\section{Collection of Manually Written PDE Control Problems}
\label{appendix:manual_data}

\begin{figure}[h!]
\vspace{-0.3em}
    \centering
\includegraphics[trim={0 154cm 0 0},clip,width=0.37\textwidth]{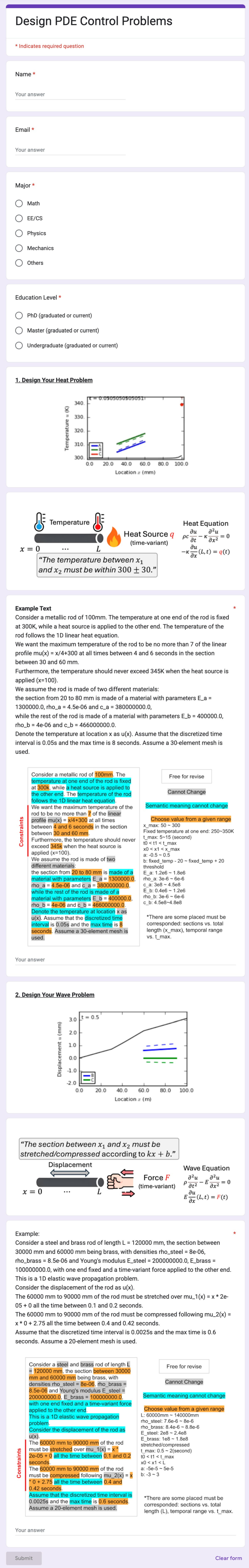}
\includegraphics[trim={0 77cm 0 48cm},clip,width=0.3\textwidth]{images/questionnaire.pdf}
\includegraphics[trim={0 4cm 0 126cm},clip,width=0.32\textwidth]{images/questionnaire.pdf}
\vspace{-0.3em}
    \caption{Google Form for collecting manually written PDE control problems.}
\vspace{-0.5em}
\label{fig:questionnaire}
\end{figure}

\begin{figure*}[h!]
    \centering
    \includegraphics[width=0.4\linewidth]{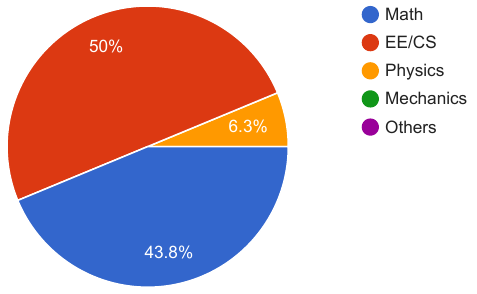}
    \includegraphics[width=0.46\linewidth]{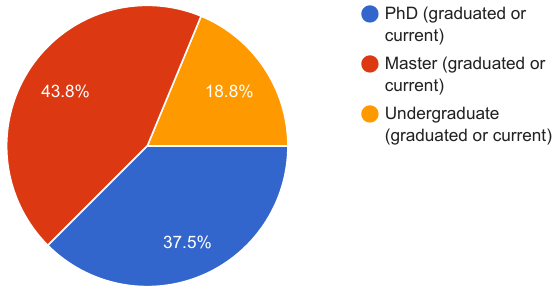}
    \caption{Background of our questionnaire participants.}
    \label{fig:questionnaire_participants}
\end{figure*}

We overview our questionnaire in Fig.~\ref{fig:questionnaire}.
For the questionnaire collection, each Zoom session takes one hour, including background instruction and manual design (with interactive guidance).
Although it is challenging to make a fair statistical comparison with human experts in terms of performance (both time cost and accuracy), our LLMs (autoformalization $\to$ Controller $\to$ code generation) complete the task in under 120 seconds on a single 6000 Ada GPU. This is significantly faster than the human volunteers we recruited (for collecting our real-world samples), who required several minutes just to read the original informal problem.
We further show the statistics of the background of participants in Fig.~\ref{fig:questionnaire_participants}.

\phantomsection\label{appendix:difference_synthetic_manual}
\paragraph{Difference between Synthetic and Manual Data}
We observe some differences between synthetic and manually generated data that may lead to the model’s failure to produce valid STL logic:
\begin{itemize}[leftmargin=*]
\item Ambiguous symbol usage, such as using “ho” instead of “rho” to denote material density.
\item Inconsistent units within a single sentence. For example, “Assume that the discretized time interval is 0.05s and the maximum time is 7400 milliseconds.”
\item Insufficient information, where four samples fail to fully describe the material properties. For instance, the manual data only provides the density of one material after the statement "the rod is composed of two materials".
\end{itemize}

\section{Training Details}
\label{appendix:training_details}

We leverage the pretrained MathCoder2-DeepSeekMath-7B~\cite{lu2024mathcoder2} checkpoint (MathCoder2) which has a 4096-token context length. All our \textit{trained} models are evaluated zero-shot. For fair comparison of our models with MathCoder2, we provide two few-shot examples for the latter to leave sufficient tokens to generate a valid output (whether STL or Python). We similarly provide two few-shot examples for GPT-4o and o1-mini.

Our instructions are structured in the Alpaca format~\cite{Taori2023alpaca}.

\subsection{Autoformalization: SFT of Translator}
The Translator was trained
with two 6000Ada GPUs using a per-GPU batch size of 16 and 4 gradient accumulation steps for a total of 3000 steps.
We fine-tuned the MathCoder2-DeepSeekMath-7B parameter model from \cite{lu2024mathcoder2} with LoRA rank $r=64$ and $\alpha = 256$.

\textbf{Prompt:} Below is a natural language description of partial differential equation optimization problem. Translate the problem into Latex code following spatial-signal temporal logic.

\subsection{Program Synthesis: SFT of Coder}
The Coder further fine-tuned the Translator with supervised fine-tuning and LoRA, rank $r=64$ and $\alpha = 256$, to produce Python code from natural language and STL pairs.
This was trained
with two 6000Ada GPUs using a per-GPU batch size of 8 and 8 gradient accumulation steps for a total of 3000 steps.

We design two prompts, firstly for generating python code aligned with the anchor STL, and secondly for generating python code aligned with the proposed subgoal STL. The natural language problem is provided in both cases to extract system settings.

\textbf{Prompt:} Below is a natural language description of partial differential equation optimization problem, paired with your spatial-signal temporal logic description of the same problem provided earlier. Note that there may be mistakes in the spatial-signal temporal logic statement but the natural language description is accurate. Translate the problem into Python code following spatial-signal temporal logic.

\textbf{Prompt:} Below is a natural language description of partial differential equation optimization problem, paired with your spatial-signal temporal logic description of an intermediate problem provided earlier. Instead of optimizing the natural language problem directly, we want to optimize the intermediate problem to produce a state that will better serve to achieve the final conditions outlined in the natural language problem. Your spatial-signal temporal logic description in latex paired to the original problem describes this intermediate problem. Translate the intermediate problem into Python code following spatial-signal temporal logic.

In practice, we find it to be helpful to supervise fine-tune the Coder with misaligned (NL, STL) pairs to promote the subgoal STL constraints when synthesizing the corresponding Python program for the subgoal STL.

\subsection{Reasoning: RLHF of Controller}
The Controller is trained with DPO~\cite{rafailov2024direct} from the Translator checkpoint with LoRA rank $r=64$ and $\alpha = 256$.
We train
with two 6000Ada GPUs using a per-GPU batch size of 2 and 4 gradient accumulation steps for a total of 16,800 steps.
For DPO, we set $\beta = 0.1$, and $\lambda = 1$ in Eq.~\ref{eq:dpo_stl}.

\textbf{Prompt:} Below is a natural language description of partial differential equation optimization problem. Instead of optimizing the provided problem directly, we want to optimize an intermediate problem to produce a state that will better serve to achieve the final conditions outlined in the natural language problem. Generate a spatial-signal temporal logic description in Latex code for such an intermediate problem.

\section{More Experiment Results}

\subsection{End-to-End Evaluation of Program Synthesis}

In Table~\ref{table:translator_coder_its_own_sstl_synthetic} and Table~\ref{table:translator_coder_its_own_sstl_human}, we provide end-to-end results of program synthesis on synthetic and manual data, respectively.
In these results, coders take LLM-generated (noisy) STLs in their prompt.
Overall, our Coder LLM still achieves strong results.
Moreover, we also show another baseline ``Coder-only'', where the Coder LLM takes only natural language as the input without explicitly formalized STLs.
In Table~\ref{table:translator_coder_its_own_sstl_human}, we can see that our method (autoformalization + program synthesis) outperforms ``Coder-only'' (direct program synthesis without autoformalization), emphasizing the importance of leveraging formal language (STL).

\begin{table}[h!]
\vspace{-0.5em}
\centering
\caption{End-to-end autoformalization and program synthesis. The Coder produces Python using the Translator's STL output. MathCoder2 produces Python using its own STL output. Coder-only is a MathCoder2 model fine-tuned to produce python directly from natural language and seeing no STL. To be comparable with the Translator + Coder autoformalization and program synthesis, Coder-only is trained for 6000 steps with the same settings as the Translator and Coder (Appendix \ref{appendix:training_details}). The evaluation for Coders is zero-shot. MathCoder2 is evaluated with two few-shot examples.
Deviations over 3 seeds are in parentheses.
\textbf{Bold} indicates the best, \ul{underline} indicates the runner up.
}
\vspace{-0.5em}

\resizebox{0.5\textwidth}{!}{
\addtolength{\tabcolsep}{-0.3em}
\begin{tabular}{ccccc}
\toprule
PDE & Model & \begin{tabular}{@{}c@{}}Executability ($\uparrow$)\\(Coder)\end{tabular} & \begin{tabular}{@{}c@{}}\textbf{Utility}\\\textbf{RMSE} ($\downarrow$)\end{tabular} \\ \midrule
\multirow{3}{*}{Heat} & Ours   & \textbf{0.9978} (0.0015) & \textbf{0.0174} (0.0065) \\
 & MathCoder2   &  0.9197 (0.02731) & 1.3841 (0.1196)  \\
 & Ours (Coder-only) &  \textbf{0.9978} (0.0009) &  \ul{0.0235} (0.00163)  \\

\midrule
\multirow{3}{*}{Wave} & Ours   & \ul{0.9620} (0.0098) & \textbf{0.0076} (0.0011) \\
 & MathCoder2   & 0.8305 (0.0475) & 0.8332 (0.0965) \\
& Ours (Coder-only) & \textbf{0.9779} (0.002148)   &	\ul{0.02522} (0.000583)  \\
\bottomrule
\end{tabular}
}
\label{table:translator_coder_its_own_sstl_synthetic}
\vspace{-0.5em}
\end{table}

\begin{table}[h!]
\vspace{-0.5em}
\centering
\caption{End-to-end autoformalization and program synthesis on \ul{manually written data} (Sec. \ref{sec:manual_data}). The Coder produces Python using the Translator's STL output. MathCoder2 produces Python using its own STL output. Coder-only is a MathCoder2 model fine-tuned to produce python directly from natural language and seeing no STL. To be comparable with the Translator + Coder autoformalization and program synthesis, Coder-only is trained for 6000 steps with the same settings as the Translator and Coder (Appendix \ref{appendix:training_details}). The evaluation for Coders is zero-shot. MathCoder2 is evaluated with two few-shot examples.
Deviations over 3 seeds are in parentheses.
\textbf{Bold} indicates the best, \ul{underline} indicates the runner up.
}
\vspace{-0.5em}

\resizebox{0.48\textwidth}{!}{
\addtolength{\tabcolsep}{-0.3em}
\begin{tabular}{ccccc}
\toprule
PDE & Model & \begin{tabular}{@{}c@{}}Executability ($\uparrow$)\\(Coder)\end{tabular} & \begin{tabular}{@{}c@{}}\textbf{Utility}\\\textbf{RMSE} ($\downarrow$)\end{tabular} \\ \midrule
\multirow{3}{*}{Heat} & Ours   & 0.4510 (0.0832) & \textbf{0.1837} (0.0095) \\
 & MathCoder2   &  \ul{0.4902} (0.1386) &  \ul{0.2928} (0.0780) \\
 & Ours (Coder-only) & \textbf{0.6078} (0.0555) & 2.442 (0.0) \\
\midrule
\multirow{3}{*}{Wave} & Ours   & \textbf{1.0} (0.0)  & \textbf{0.0119} (0.0)  \\
 & MathCoder2   & \ul{0.9020} (0.0277)  & 1.767 (0.8109) \\
 & Ours (Coder-only) & 0.4706 (0.0)  & \ul{0.0890} (0.0)  \\
\bottomrule
\end{tabular}
}
\label{table:translator_coder_its_own_sstl_human}
\vspace{-0.5em}
\end{table}

\subsection{PDE Reasoning}

\subsubsection{End-to-End Evaluation of PDE Reasoning}
We include end-to-end evaluation results, where we use the Coder LLM to generate Python code after the Controller LLM propose subgoal STLs in Table~\ref{table:controller_end2end}. 
In general, our Controller still shows strong reasoning capability (both the success rate and utility gain), and also high rate of proposing valid subgoal STL ($\phi^\prime$).

\begin{table*}[t!]
\centering
\caption{
End-to-end scientific reasoning over PDE control problems via our Controller LLM.
Deviations over 5 seeds are in parentheses.
``Valid STL $\phi^\prime$ (\%)'' is the ratio of proposed subgoal STL $\phi^\prime$ without any syntax errors.
\textbf{Bold} indicates the best, \ul{underline} denotes the runner-up. 
“x” indicates no valid STLs were generated for evaluation. “-” indicates not applicable.}
\vspace{-0.5em}
\resizebox{0.99\textwidth}{!}{
\addtolength{\tabcolsep}{-0.3em}
\begin{tabular}{cc|cccc|cccc}
\toprule
\multirow{3}{*}{PDE} & \multirow{3}{*}{\begin{tabular}{@{}c@{}}Difficulty\\Level\end{tabular}} & \multicolumn{4}{c|}{Success Rate $\overline{P}$ $\uparrow$} & \multicolumn{4}{c}{Utility Gain $\overline{\Delta r}$ $\uparrow$} \\ \cmidrule{3-6} \cmidrule{7-10}
 &  & Ours & \begin{tabular}[c]{@{}c@{}}Math-\\Coderv2\end{tabular} & \begin{tabular}[c]{@{}c@{}}GPT\\(o1-mini)\end{tabular} & \begin{tabular}[c]{@{}c@{}}GPT\\(4o)\end{tabular} & Ours & \begin{tabular}[c]{@{}c@{}}Math-\\Coderv2\end{tabular} & \begin{tabular}[c]{@{}c@{}}GPT\\(o1-mini)\end{tabular} & \begin{tabular}[c]{@{}c@{}}GPT\\(4o)\end{tabular} \\ \midrule
\multirow{5}{*}{Heat} 
 & Easy & \textbf{0.490} (0.1666) & \ul{0.399} (0.2605) & 0.154 (0) & 0.160 (0.0686) & \textbf{0.651} (1.1150) & \ul{0.352} (1.3309) & -1.262 (0) & -0.865 (2.0310)  \\ \cmidrule{2-10}
 & Medium & \textbf{0.345} (0.1640) & \ul{0.339} (0.2345) & 0 (0) & 0.121 (0.0497) & \ul{-1.055} (0.9657) & \textbf{-0.061} (0.4098) & -2.455 (0) & -2.294 (1.7773)  \\ \cmidrule{2-10}
 & Hard & \textbf{0.307} (0.1363)  & \ul{0.239} (0.1895) & 0.074 (0.0262) & 0.150 (0.0797) & \ul{-1.788} (2.1369) & \textbf{-1.254} (0.8047) & -3.992 (0.5175) & -2.603 (2.4486)  \\ \cmidrule{2-10}
 & All & \textbf{0.381} (0.1556) & \ul{0.326} (0.2282) & x & 0.144 (0.0660) & \ul{-0.731} (1.4059) & \textbf{-0.321} (0.8485) & x & -1.921 (2.0856)  \\ \midrule
\multirow{5}{*}{Wave} 
 & Easy & \ul{0.897} (0.0502) & 0.854 (0.0894) & x & \textbf{1} (0) & \ul{1.423} (0.4135) & 1.173 (0.6424) & x & \textbf{1.907} (0.0333)  \\ \cmidrule{2-10}
 & Medium & \ul{0.836} (0.0936) & 0.743 (0.0410) & x & \textbf{0.958} (0.0105) & \textbf{0.865} (0.4508) & 0.758 (0.4265) & x & \ul{0.803} (0.2279)  \\ \cmidrule{2-10}
 & Hard & 0.331 (0.1214) & \ul{0.336} (0.1041) & x & \textbf{0.416} (0.1113) & \textbf{-0.289} (0.5723) & -0.649 (0.3948) & x & \ul{-0.376} (0.2142) \\ \cmidrule{2-10}
 & All & \ul{0.688} (0.0884) & 0.644 (0.0781) & x & \textbf{0.791} (0.0406) & \ul{0.698} (0.5773) & 0.427 (0.4879) & x & \textbf{0.778} (0.1585)  \\ \midrule \midrule
 \multicolumn{2}{c|}{Valid STL $\phi^\prime$ (\%) ($\uparrow$)} & \textbf{82.70} (1.97) & \ul{42.45} (10.54) & 0.04 (0.10) & 2.55 (0.65) & \textbf{82.70} (1.97) & \ul{42.45} (10.54) & 0.04 (0.10) & 2.55 (0.65)\\
 \midrule
 \multicolumn{2}{c|}{Valid Python Code (\%) ($\uparrow$) } & \textbf{75.65} (1.50) & \ul{27.95} (5.72) & 0.09 (0.78) & 3.25 (0.63) & \textbf{75.65} (1.50) & \ul{27.95} (5.72) & 0.09 (0.78) & 3.25 (0.63)\\
 \bottomrule
\end{tabular}
}
\label{table:controller_end2end}
\vspace{-0.5em}
\end{table*}

\subsubsection{Proportion of valid predictions}
We notice that the number of valid predictions varied significantly depending on the model. Therefore, we include Table~\ref{table:exec_stl} for STL and Table~\ref{table:exec_python} for Python program to comprehensively show the number of valid predictions that each model makes under each type of problem and difficulty level. Please note that the values in Table~\ref{table:exec_stl} and Table~\ref{table:exec_python} are not expressed as percentages.

\begin{table*}[t!]
\centering
\caption{
Proportion of valid subgoal STL generations by different controller models in Table~\ref{table:controller} and~\ref{table:controller_end2end} (in decimal form). Deviations over 5 seeds are in parentheses.}
\vspace{-0.5em}
\resizebox{0.55\textwidth}{!}{
\addtolength{\tabcolsep}{-0.3em}
\begin{tabular}{cc|cccc}
\toprule
\multirow{3}{*}{PDE} & \multirow{3}{*}{\begin{tabular}{@{}c@{}}Difficulty\\Level\end{tabular}} & \multicolumn{4}{c}{Proportion ($\in [0, 1]$) of valid STL $\phi^\prime$ ($\uparrow$)} \\ \cmidrule{3-6}
 &  & Ours & \begin{tabular}[c]{@{}c@{}}Math-\\Coderv2\end{tabular} & \begin{tabular}[c]{@{}c@{}}GPT\\(o1-mini)\end{tabular} & \begin{tabular}[c]{@{}c@{}}GPT\\(4o)\end{tabular} \\ \midrule
\multirow{5}{*}{Heat} 
 & Easy & 0.8851 (0.0110) & 0.507 (0.1354) & 0 & 0.028 (0.0078) \\ \cmidrule{2-6}
 & Medium & 0.891 (0.0243) & 0.491 (0.0962) & 0.0004 (0.0015) & 0.031 (0.0046)  \\ \cmidrule{2-6}
 & Hard & 0.881 (0.0148) & 0.493 (0.0956) & 0.002 (0.0045) & 0.052 (0.0113)  \\ \cmidrule{2-6}
 & All & 0.886 (0.0167) & 0.497 (0.1091) & 0.0008 (0.002) & 0.037 (0.0079)  \\ \midrule
\multirow{5}{*}{Wave} 
 & Easy & 0.756 (0.0228) & 0.309 (0.1113) & 0 & 0.008 (0.0066)  \\ \cmidrule{2-6}
 & Medium & 0.748 (0.0257) & 0.397 (0.0947) & 0 & 0.015 (0.0052)  \\ \cmidrule{2-6}
 & Hard & 0.799 (0.0197) & 0.349 (0.0990) & 0 & 0.018 (0.0036) \\ \cmidrule{2-6}
 & All & 0.768 (0.0227) & 0.352 (0.1016) & 0 & 0.014 (0.0051)  \\ \bottomrule
\end{tabular}
}
\label{table:exec_stl}
\vspace{-0.5em}
\end{table*}

\begin{table*}[t!]
\centering
\caption{
Proportion of valid subgoal Python program generations by different coder models in Table~\ref{table:controller_end2end} (in decimal form). Deviations over 5 seeds are in parentheses.}
\vspace{-0.5em}
\resizebox{0.55\textwidth}{!}{
\addtolength{\tabcolsep}{-0.3em}
\begin{tabular}{cc|cccc}
\toprule
\multirow{3}{*}{PDE} & \multirow{3}{*}{\begin{tabular}{@{}c@{}}Difficulty\\Level\end{tabular}} & \multicolumn{4}{c}{Proportion ($\in [0, 1]$) of valid Python program ($\uparrow$)} \\ \cmidrule{3-6}
 &  & Ours & \begin{tabular}[c]{@{}c@{}}Math-\\Coderv2\end{tabular} & \begin{tabular}[c]{@{}c@{}}GPT\\(o1-mini)\end{tabular} & \begin{tabular}[c]{@{}c@{}}GPT\\(4o)\end{tabular} \\ \midrule
\multirow{5}{*}{Heat} 
 & Easy & 0.754 (0.0244)  & 0.301 (0.0624) & 0.014 (0.0193) & 0.047 (0.0113) \\ \cmidrule{2-6}
 & Medium & 0.767 (0.0261)  & 0.295 (0.0503) & 0.012 (0.0124) & 0.039 (0.0048)  \\ \cmidrule{2-6}
 & Hard & 0.828 (0.0113)  & 0.329 (0.0532) & 0.027 (0.0147) & 0.058 (0.0085)  \\ \cmidrule{2-6}
 & All & 0.783 (0.0206)  & 0.308 (0.0553) & 0.018 (0.0155) & 0.048 (0.0082)  \\ \midrule
\multirow{5}{*}{Wave} 
 & Easy & 0.702 (0.0080)  & 0.211 (0.0675) & 0 & 0.012 (0.0048)  \\ \cmidrule{2-6}
 & Medium & 0.713 (0.0113)  & 0.278 (0.0538) & 0 & 0.018 (0.0041)  \\ \cmidrule{2-6}
 & Hard & 0.776 (0.0086)  & 0.265 (0.0561) & 0 & 0.021 (0.0044) \\ \cmidrule{2-6}
 & All & 0.730 (0.0093) & 0.251 (0.0591) & 0 & 0.017 (0.0044)  \\ \bottomrule
\end{tabular}
}
\label{table:exec_python}
\vspace{-0.5em}
\end{table*}

\section{More Related Works}
\label{appendix:related works}

\vspace{-0.5em}
\subsection{Autoformalization in AI-for-math}

Autoformalization, the process of converting informal mathematical statements or instructions into formal representations, is explored through a variety of techniques.
A significant subset of works employed LLMs to translate informal descriptions into formal representations. 
VernaCopter~\cite{van2024vernacopter} leveraged LLMs to convert natural language commands into Signal Temporal Logic (STL) specifications, integrating syntax and semantic checkers for correctness. Pretrained LLMs like GPT-3.5 and GPT-4o were leveraged to translate informal problems into Isabelle proof sketches, refining outputs through iterative prompting and heuristic-based validation~\cite{zhou2024don}. Back-translation~\cite{jiang2023multilingual} trained LLMs to map between informal and formal theorem statements in Lean4 and Isabelle. These approaches focused on leveraging LLMs for direct autoformalization while incorporating filtering mechanisms to improve reliability.
In contrast, hybrid approaches interact between manual and autoformalization. Several studies combined expert-curated manual formalization with automated techniques to improve accuracy. DeepSeek-Prover~\cite{xin2024deepseek}, Trigo~\cite{xiong2023trigo}, and~\cite{murphy2024autoformalizing} used an iterative pipeline where initial formalization was manually crafted, followed by automated expansion and refinement. LILA~\cite{mishra2022lila} similarly applied domain-specific rules and Python-based DSL annotations for automatic formalization while relying on human annotators for cases where automation failed. These hybrid approaches aimed to balance the scalability of automation with the precision of manual verification.
In our work, we for the first time train LLMs to autoformalize informal PDE control problems into formalized STL logic.

\vspace{-0.5em}
\subsection{PDE Controls}

PDEs are essential for modeling physical phenomena, helping researchers predict behaviors, optimize processes, and drive innovation across fields like climate modeling and material design.
PDE control focuses on manipulating system behaviors, ensuring stability in applications like robotics and reactors, while enabling systems to adapt to changing conditions for more sustainable solutions
~\cite{alvarez2020formal,holl2020learning,ramos2022control,mukherjee2023actor,wei2024generative,wei2024closed}.
Adjoint methods have been widely used for controlling physical systems governed by PDEs due to their accuracy, despite being computationally expensive~\cite{lions1971optimal,mcnamara2004fluid,protas2008adjoint}, while deep learning-based approaches, such as supervised learning with differentiable physics losses~\cite{holl2020learning,hwang2022solving}, optimize control directly via backpropagation through time.
\cite{mowlavi2022optimalcontrolpdesusing,barrystraume2022physicsinformedneuralnetworkspdeconstrained} optimizes control problems and the PDE system state with physics-informed neural networks.
Reinforcement learning (RL)~\cite{farahmand2017deep, pan2018reinforcement, rabault2019artificial} optimizes control by treating signals as actions for sequential decision-making in fluid dynamics applications like drag reduction, heat transfer, and swimming \cite{garnier2021review}.
Pretrained LLMs can enhance PDE surrogate modeling by integrating textual descriptions of system information--such as boundary conditions and the governing PDE--into a multimodal learning framework~\cite{lorsung2025explainlikeimfive}.
PDE control can also be discretized and formulated into a mixed-integer linear programming (MILP) problem via finite element method (FEM)~\cite{sadraddini2015robust,alvarez2020formal}.
More recently, diffusion-based generation has been leveraged to jointly optimize the PDE simulations and control signals~\cite{wei2024diffphycon}.
Different from traditional PDE control objectives such as setpoint/trajectory tracking and disturbance rejection~\cite{paunonen2019reduced,paunonen2022robust}, our utility score objective (Appendix~\ref{appendix:utility_stl}), which mainly aims to reduce distance to the target, can better handle inequality constraints compared to tracking errors.
From an optimization perspective, our work leverages LLMs to propose better initializations (initial conditions) to solve the open-loop PDE control problem. Extensions to closed-loop control are possible by appending the utility from the LLM-proposed optimization into future optimization rounds. This increases the complexity of LLM fine-tuning; thus, as the first step in this direction we focus on open-loop control.

\vspace{-0.5em}
\subsection{LLM-based Task Planning}

Natural language (NL)-based task planning in robotics has gained increasing attention. Approaches such as~\cite{pan2023dataefficientlearningnaturallanguage} enable task-specific translations from informal language to Linear Temporal Logic (LTL), allowing robots to follow structured plans even in low-resource scenarios.
Building on this foundation, recent research has explored the use of LLMs for task planning, demonstrating models' potential in decision-making and executing complex plans~\cite{singh2023progprompt, shah2023lm, li2022pre}.
For instance,~\cite{wangconformal} leveraged LLMs to arrange and predict execution sequences for robots, achieving a comparable success rate to human users.
Additionally, \cite{ren2023robots} addressed the hallucination issue from LLM-based planners by incorporating uncertainty alignment, improving the reliability of generated plans.
More recently, CLMASP~\cite{lin2024clmasp} refined LLM-generated skeleton plans using Answer Set Programming (ASP) for robotic task execution.
Inspired by these subgoal approaches, we for the first time train LLMs to perform reasoning and planning on PDE control problems.

\section{Datasheets for Datasets}

This document is based on \textit{Datasheets for Datasets} by~\cite{gebru2021datasheets}.

\subsection{Motivation}

    \textcolor{\sectioncolor}{\textbf{For what purpose was the dataset created?
    }
    Was there a specific task in mind? Was there
    a specific gap that needed to be filled? Please provide a description.
    } \\
    The dataset was created to enable large language models (LLMs) to tackle complex Partial Differential Equation (PDE) control problems. The specific purpose is to advance automated formalization and reasoning in applied mathematics, addressing the lack of datasets tailored to PDE-related tasks. The dataset bridges informal natural language problems and formal specifications/code for PDE systems, fostering research in scientific computing and engineering.
    \\
    
    \textcolor{\sectioncolor}{\textbf{Who created this dataset (e.g., which team, research group) and on behalf
    of which entity (e.g., company, institution, organization)?
    }
    } \\
    The dataset was created by the anonymous authors of this PDE-Controller paper, affiliated with a research group focused on AI-for-math applications. \\
    
    \textcolor{\sectioncolor}{\textbf{What support was needed to make this dataset?
    }
    (e.g.who funded the creation of the dataset? If there is an associated
    grant, provide the name of the grantor and the grant name and number, or if
    it was supported by a company or government agency, give those details.)
    } \\
    The creation of the dataset was supported by research funding for developing novel applications of LLMs in applied mathematics. Further support included computational resources for fine-tuning LLMs and manual curation by domain experts. \\
    
    \textcolor{\sectioncolor}{\textbf{Any other comments?
    }} \\
    The dataset represents a pioneering effort to merge AI capabilities with PDE-based scientific reasoning, significantly expanding the potential applications of LLMs. \\

\subsection{Composition}
    \textcolor{\sectioncolor}{\textbf{What do the instances that comprise the dataset represent (e.g., documents,
    photos, people, countries)?
    }
    Are there multiple types of instances (e.g., movies, users, and ratings;
    people and interactions between them; nodes and edges)? Please provide a
    description.
    } \\
    Each instance represents a PDE control problem, including:
1) Informal problem descriptions in natural language;
2) Formal specifications in Signal Temporal Logic (STL);
3) Python code that integrates PDE simulation and optimization tools. \\
    
    \textcolor{\sectioncolor}{\textbf{How many instances are there in total (of each type, if appropriate)?
    }
    } \\
    The dataset comprises over 2.13 million synthetic (natural language, STL, Python code) triplets, with additional real-world examples including 17 manually written heat problems and 17 wave problems.
    \\
    
    \textcolor{\sectioncolor}{\textbf{Does the dataset contain all possible instances or is it a sample (not
    necessarily random) of instances from a larger set?
    }
    If the dataset is a sample, then what is the larger set? Is the sample
    representative of the larger set (e.g., geographic coverage)? If so, please
    describe how this representativeness was validated/verified. If it is not
    representative of the larger set, please describe why not (e.g., to cover a
    more diverse range of instances, because instances were withheld or
    unavailable).
    } \\
    It is a synthesized dataset designed to cover a diverse range of PDE control problems, sampled and augmented from representative templates to ensure coverage of key scenarios and complexities. \\
    
    \textcolor{\sectioncolor}{\textbf{What data does each instance consist of?
    }
    “Raw” data (e.g., unprocessed text or images) or features? In either case,
    please provide a description.
    } \\
    Each instance includes:
1) Informal natural language descriptions of PDE problems;
2) Formal representations in STL syntax;
3) Python code for solving the PDE problem using optimizers such as Gurobi.
    
    \textcolor{\sectioncolor}{\textbf{Is there a label or target associated with each instance?
    }
    If so, please provide a description.
    } \\
    Yes, each instance includes ground-truth STL and Python code, verified for alignment and executability. \\
    
    \textcolor{\sectioncolor}{\textbf{Is any information missing from individual instances?
    }
    If so, please provide a description, explaining why this information is
    missing (e.g., because it was unavailable). This does not include
    intentionally removed information, but might include, e.g., redacted text.
    } \\
    Not Applicable. \\
    
    \textcolor{\sectioncolor}{\textbf{Are relationships between individual instances made explicit (e.g., users’
    movie ratings, social network links)?
    }
    If so, please describe how these relationships are made explicit.
    } \\
    Yes, relationships between natural language, STL specifications, and Python code are explicitly maintained for traceability. \\
    
    \textcolor{\sectioncolor}{\textbf{Are there recommended data splits (e.g., training, development/validation,
    testing)?
    }
    If so, please provide a description of these splits, explaining the
    rationale behind them.
    } \\
    Yes, the dataset is split into training and testing sets, with specific splits for heat and wave problems. \\
    
    \textcolor{\sectioncolor}{\textbf{Are there any errors, sources of noise, or redundancies in the dataset?
    }
    If so, please provide a description.
    } \\
    Synthetic data is validated through automated checks and human verification. Errors may arise from annotation inconsistencies, especially in manually curated problems. \\
    
    \textcolor{\sectioncolor}{\textbf{Is the dataset self-contained, or does it link to or otherwise rely on
    external resources (e.g., websites, tweets, other datasets)?
    }
    If it links to or relies on external resources, a) are there guarantees
    that they will exist, and remain constant, over time; b) are there official
    archival versions of the complete dataset (i.e., including the external
    resources as they existed at the time the dataset was created); c) are
    there any restrictions (e.g., licenses, fees) associated with any of the
    external resources that might apply to a future user? Please provide
    descriptions of all external resources and any restrictions associated with
    them, as well as links or other access points, as appropriate.
    } \\
    The dataset is self-contained, with no reliance on external or dynamic resources. \\
    
    \textcolor{\sectioncolor}{\textbf{Does the dataset contain data that might be considered confidential (e.g.,
    data that is protected by legal privilege or by doctor-patient
    confidentiality, data that includes the content of individuals’ non-public
    communications)?
    }
    If so, please provide a description.
    } \\
    No. \\
    
    \textcolor{\sectioncolor}{\textbf{Does the dataset contain data that, if viewed directly, might be offensive,
    insulting, threatening, or might otherwise cause anxiety?
    }
    If so, please describe why.
    } \\
    No. \\
    
    \textcolor{\sectioncolor}{\textbf{Does the dataset relate to people?
    }
    If not, you may skip the remaining questions in this section.
    } \\
    No. \\
    
    \textcolor{\sectioncolor}{\textbf{Does the dataset identify any subpopulations (e.g., by age, gender)?
    }
    If so, please describe how these subpopulations are identified and
    provide a description of their respective distributions within the dataset.
    } \\
    No. \\
    
    \textcolor{\sectioncolor}{\textbf{Is it possible to identify individuals (i.e., one or more natural persons),
    either directly or indirectly (i.e., in combination with other data) from
    the dataset?
    }
    If so, please describe how.
    } \\
    No. \\
    
    \textcolor{\sectioncolor}{\textbf{Does the dataset contain data that might be considered sensitive in any way
    (e.g., data that reveals racial or ethnic origins, sexual orientations,
    religious beliefs, political opinions or union memberships, or locations;
    financial or health data; biometric or genetic data; forms of government
    identification, such as social security numbers; criminal history)?
    }
    If so, please provide a description.
    } \\
    No. \\
    
    \textcolor{\sectioncolor}{\textbf{Any other comments?
    }} \\
    The dataset’s richness in complexity and diversity makes it a significant resource for advancing applied mathematics via AI. \\

\subsection{Collection}

    \textcolor{\sectioncolor}{\textbf{How was the data associated with each instance acquired?
    }
    Was the data directly observable (e.g., raw text, movie ratings),
    reported by subjects (e.g., survey responses), or indirectly
    inferred/derived from other data (e.g., part-of-speech tags, model-based
    guesses for age or language)? If data was reported by subjects or
    indirectly inferred/derived from other data, was the data
    validated/verified? If so, please describe how.
    } \\
    The data was synthesized from key PDE control templates, augmented through principled methods, and verified by experts. Real-world problems were collected via a questionnaire-based manual curation process involving students and researchers. \\
    
    \textcolor{\sectioncolor}{\textbf{Over what timeframe was the data collected?
    }
    Does this timeframe match the creation timeframe of the data associated
    with the instances (e.g., recent crawl of old news articles)? If not,
    please describe the timeframe in which the data associated with the
    instances was created. Finally, list when the dataset was first published.
    } \\
    The synthetic dataset was generated in late 2024, with real-world problems curated in 2025. \\
    
    \textcolor{\sectioncolor}{\textbf{What mechanisms or procedures were used to collect the data (e.g., hardware
    apparatus or sensor, manual human curation, software program, software
    API)?
    }
    How were these mechanisms or procedures validated?
    } \\
    Procedures included automated STL generation, natural language augmentation using GPT-4, and manual problem formulation. \\
    
    \textcolor{\sectioncolor}{\textbf{What was the resource cost of collecting the data?
    }
    (e.g. what were the required computational resources, and the associated
    financial costs, and energy consumption - estimate the carbon footprint.)} \\
    Resource costs included computational expenses for data synthesis and human time for manual curation and validation.
    \\
    
    \textcolor{\sectioncolor}{\textbf{If the dataset is a sample from a larger set, what was the sampling
    strategy (e.g., deterministic, probabilistic with specific sampling
    probabilities)?
    }
    } \\
    Not applicable. \\
    
    \textcolor{\sectioncolor}{\textbf{Who was involved in the data collection process (e.g., students,
    crowdworkers, contractors) and how were they compensated (e.g., how much
    were crowdworkers paid)?
    }
    } \\
    Graduate students and researchers with expertise in applied mathematics and AI. \\
    
    \textcolor{\sectioncolor}{\textbf{Were any ethical review processes conducted (e.g., by an institutional
    review board)?
    }
    If so, please provide a description of these review processes, including
    the outcomes, as well as a link or other access point to any supporting
    documentation.
    } \\
    No. \\
    
    \textcolor{\sectioncolor}{\textbf{Does the dataset relate to people?
    }
    If not, you may skip the remainder of the questions in this section.
    } \\
    No. \\
    
    \textcolor{\sectioncolor}{\textbf{Did you collect the data from the individuals in question directly, or
    obtain it via third parties or other sources (e.g., websites)?
    }
    } \\
    Our manually written data is collected from each individual with questions.\\
    
    \textcolor{\sectioncolor}{\textbf{Were the individuals in question notified about the data collection?
    }
    If so, please describe (or show with screenshots or other information) how
    notice was provided, and provide a link or other access point to, or
    otherwise reproduce, the exact language of the notification itself.
    } \\
    N/A \\
    
    \textcolor{\sectioncolor}{\textbf{Did the individuals in question consent to the collection and use of their
    data?
    }
    If so, please describe (or show with screenshots or other information) how
    consent was requested and provided, and provide a link or other access
    point to, or otherwise reproduce, the exact language to which the
    individuals consented.
    } \\
    N/A \\
    
    \textcolor{\sectioncolor}{\textbf{If consent was obtained, were the consenting individuals provided with a
    mechanism to revoke their consent in the future or for certain uses?
    }
     If so, please provide a description, as well as a link or other access
     point to the mechanism (if appropriate)
    } \\
    N/A \\
    
    \textcolor{\sectioncolor}{\textbf{Has an analysis of the potential impact of the dataset and its use on data
    subjects (e.g., a data protection impact analysis)been conducted?
    }
    If so, please provide a description of this analysis, including the
    outcomes, as well as a link or other access point to any supporting
    documentation.
    } \\
    No. Our data are intended to be used in evaluation only and all charts are publicly avialable.  \\
    
    \textcolor{\sectioncolor}{\textbf{Any other comments?
    }} N/A \\

\subsection{Preprocessing / Cleaning / Labeling}

    \textcolor{\sectioncolor}{\textbf{Was any preprocessing/cleaning/labeling of the data
    done(e.g.,discretization or bucketing, tokenization, part-of-speech
    tagging, SIFT feature extraction, removal of instances, processing of
    missing values)?
    }
    If so, please provide a description. If not, you may skip the remainder of
    the questions in this section.
    } \\
    Yes, preprocessing included:
1) Reformatting natural language prompts;
2) Validating STL and Python code for correctness and executability;
3) Augmenting natural language data using rephrasing techniques. \\

    \textcolor{\sectioncolor}{\textbf{Was the “raw” data saved in addition to the preprocessed/cleaned/labeled
    data (e.g., to support unanticipated future uses)?
    }
    If so, please provide a link or other access point to the “raw” data.
    } \\
    Yes, raw data and intermediate representations are retained for reproducibility and future use. \\

    \textcolor{\sectioncolor}{\textbf{Is the software used to preprocess/clean/label the instances available?
    }
    If so, please provide a link or other access point.
    } \\
    The tools and scripts for preprocessing are included in the supplementary materials of the PDE-Controller framework. \\

    \textcolor{\sectioncolor}{\textbf{Any other comments?
    }} \\
    Preprocessing ensures high-quality alignment between natural language, formal logic, and executable code. \\

\subsection{Uses}

    \textcolor{\sectioncolor}{\textbf{Has the dataset been used for any tasks already?
    }
    If so, please provide a description.
    } \\
    Yes, it was used to train and evaluate the PDE-Controller framework and benchmark its performance against state-of-the-art LLMs. \\

    \textcolor{\sectioncolor}{\textbf{Is there a repository that links to any or all papers or systems that use the dataset?
    }
    If so, please provide a link or other access point.
    } \\
    N/A \\

    \textcolor{\sectioncolor}{\textbf{What (other) tasks could the dataset be used for?
    }
    } \\
    The dataset could be used for:
1) Training models for scientific reasoning and formalization;
2) Developing tools for automated program synthesis;
3) Advancing research in AI-driven engineering and physics. \\

    \textcolor{\sectioncolor}{\textbf{Is there anything about the composition of the dataset or the way it was
    collected and preprocessed/cleaned/labeled that might impact future uses?
    }
    For example, is there anything that a future user might need to know to
    avoid uses that could result in unfair treatment of individuals or groups
    (e.g., stereotyping, quality of service issues) or other undesirable harms
    (e.g., financial harms, legal risks) If so, please provide a description.
    Is there anything a future user could do to mitigate these undesirable
    harms?
    } \\
    N/A \\

    \textcolor{\sectioncolor}{\textbf{Are there tasks for which the dataset should not be used?
    }
    If so, please provide a description.
    } \\
    The dataset is not suitable for tasks unrelated to PDE control or tasks requiring real-world human data. \\

    \textcolor{\sectioncolor}{\textbf{Any other comments?
    }} \\
    The dataset's structured format supports reproducible and extensible research in applied mathematics. \\

\subsection{Distribution}

    \textcolor{\sectioncolor}{\textbf{Will the dataset be distributed to third parties outside of the entity
    (e.g., company, institution, organization) on behalf of which the dataset
    was created?
    }
    If so, please provide a description.
    } \\
    Yes, the dataset will be made publicly available for research purposes. \\

    \textcolor{\sectioncolor}{\textbf{How will the dataset will be distributed (e.g., tarball on website, API,
    GitHub)?
    }
    Does the dataset have a digital object identifier (DOI)?
    } \\
    The dataset will be distributed via GitHub and academic repositories, with accompanying documentation. \\

    \textcolor{\sectioncolor}{\textbf{When will the dataset be distributed?
    }
    } \\
    The dataset is expected to be released following the ICML 2025 conference. \\

    \textcolor{\sectioncolor}{\textbf{Will the dataset be distributed under a copyright or other intellectual
    property (IP) license, and/or under applicable terms of use (ToU)?
    }
    If so, please describe this license and/or ToU, and provide a link or other
    access point to, or otherwise reproduce, any relevant licensing terms or
    ToU, as well as any fees associated with these restrictions.
    } \\
    Yes, it will be distributed under a permissive license (e.g., CC BY-SA 4.0) to encourage research use. \\

    \textcolor{\sectioncolor}{\textbf{Have any third parties imposed IP-based or other restrictions on the data
    associated with the instances?
    }
    If so, please describe these restrictions, and provide a link or other
    access point to, or otherwise reproduce, any relevant licensing terms, as
    well as any fees associated with these restrictions.
    } \\
    All charts are subjected to their respective copyrights by the authors of this paper. \\

    \textcolor{\sectioncolor}{\textbf{Do any export controls or other regulatory restrictions apply to the
    dataset or to individual instances?
    }
    If so, please describe these restrictions, and provide a link or other
    access point to, or otherwise reproduce, any supporting documentation.
    } \\
    N/A \\

    \textcolor{\sectioncolor}{\textbf{Any other comments?
    }} \\
    Distribution will include detailed usage guidelines to ensure proper application of the dataset. \\

\subsection{Maintenance}

    \textcolor{\sectioncolor}{\textbf{Who is supporting/hosting/maintaining the dataset?
    }
    } \\
    The authors of the PDE-Controller framework. \\

    \textcolor{\sectioncolor}{\textbf{How can the owner/curator/manager of the dataset be contacted (e.g., email
    address)?
    }
    } \\
    Contact information will be provided with the dataset release. \\

    \textcolor{\sectioncolor}{\textbf{Is there an erratum?
    }
    If so, please provide a link or other access point.
    } \\
    \url{https://pde-controller.github.io/} \\

    \textcolor{\sectioncolor}{\textbf{Will the dataset be updated (e.g., to correct labeling errors, add new
    instances, delete instances)?
    }
    If so, please describe how often, by whom, and how updates will be
    communicated to users (e.g., mailing list, GitHub)?
    } \\
    Yes, periodic updates will incorporate additional real-world problems and refinements. \\

    \textcolor{\sectioncolor}{\textbf{If the dataset relates to people, are there applicable limits on the
    retention of the data associated with the instances (e.g., were individuals
    in question told that their data would be retained for a fixed period of
    time and then deleted)?
    }
    If so, please describe these limits and explain how they will be enforced.
    } \\
    N/A \\

    \textcolor{\sectioncolor}{\textbf{Will older versions of the dataset continue to be
    supported/hosted/maintained?
    }
    If so, please describe how. If not, please describe how its obsolescence
    will be communicated to users.
    } \\
    Yes, previous versions will remain accessible for reproducibility. \\

    \textcolor{\sectioncolor}{\textbf{If others want to extend/augment/build on/contribute to the dataset, is
    there a mechanism for them to do so?
    }
    If so, please provide a description. Will these contributions be
    validated/verified? If so, please describe how. If not, why not? Is there a
    process for communicating/distributing these contributions to other users?
    If so, please provide a description.
    } \\
    Yes, contributions will be encouraged through a collaborative platform (e.g., GitHub).\\

    \textcolor{\sectioncolor}{\textbf{Any other comments?
    }} \\
    The dataset’s maintainers are committed to ensuring its long-term usability and relevance for scientific research.

\vspace{-2ex}
\section{Misc.}
\textbf{URL to benchmark.} The benchmark URL can be found here: 
\url{https://pde-controller.github.io/}

\textbf{URL to Croissant metadata.} The Croissant metadata URL can be found here:
\url{https://huggingface.co/datasets/delta-lab-ai/pde-controller/tree/main}

\textbf{Author statement \& license information.} We the authors bear all responsibility in case of violation of rights.

\textbf{Hosting and maintenance.}
We have a dedicated webpage for hosting instructions: \url{https://pde-controller.github.io/}.
We are committed to performing major maintenance periodically.

\textbf{Dataset Structure.}
All files are stored in the JSONL format. For the translator dataset, we store separate files based on STL syntax formats, the number of constraints, and the train-test split. Each training file contains more than 600 samples, and each test file contains more than 150 samples. Each sample includes the informal question in natural language, the STL representation in LaTeX, and the corresponding Python script.

For the preference dataset, we split the files based on three difficulty levels and the train-test split. Each sample in the file contains the informal question in natural language, the winner STL that improves the informal question, and the loser STL that worsens it. For each STL, we also provide the resulting utility score.

\end{document}